%% file: neurips_2026.tex
\documentclass{article}

\PassOptionsToPackage{numbers, compress}{natbib}

\usepackage[main, final]{neurips_2026}

\usepackage[utf8]{inputenc} 
\usepackage[T1]{fontenc}    
\usepackage{hyperref}
\usepackage{url}            
\usepackage{booktabs}       
\usepackage{amsfonts}       
\usepackage{nicefrac}       
\usepackage{microtype}      
\usepackage{xcolor}         

\usepackage{amsmath}        
\usepackage{amssymb}        
\usepackage{dsfont}
\usepackage{amsthm}         
\usepackage{mathtools}      
\usepackage{graphicx}
\usepackage{subcaption}
\usepackage{wrapfig}
\usepackage{algorithm}      
\usepackage{algpseudocode}  
\usepackage{enumitem}       
\usepackage{tabularx}
\usepackage{array}
\usepackage{etoc}
\usepackage{colortbl}
\usepackage{multirow}
\usepackage{pifont}
\usepackage{tikz}
\usetikzlibrary{fit, shapes, arrows.meta, positioning}
\usepackage[breakable]{tcolorbox}
\tcbuselibrary{listings, skins}
\usepackage{listings}
\definecolor{pbsage}{HTML}{e8f1ec}
\definecolor{pbmint}{HTML}{eef5f3}
\definecolor{bandhdr}{HTML}{f5f9f6}
\definecolor{bestcell}{HTML}{ffd0c0}
\definecolor{secondcell}{HTML}{fff5d0}
\definecolor{updelta}{HTML}{1f6f3f}
\definecolor{downdelta}{HTML}{a6242c}
\definecolor{contentColor}{HTML}{1f4e79}
\definecolor{logicColor}{HTML}{6a1b9a}
\newcommand{\contentP}[1]{\textcolor{contentColor}{\emph{#1}}}
\newcommand{\logicP}[1]{\textcolor{logicColor}{\emph{#1}}}
\definecolor{majGray}{HTML}{ededed}
\definecolor{subA}{HTML}{f6efde}
\definecolor{subB}{HTML}{f5e1d8}
\definecolor{subC}{HTML}{e7eef7}
\definecolor{subD}{HTML}{e9e0ee}
\definecolor{subE}{HTML}{d8e6ec}
\definecolor{ablA}{HTML}{ededed}
\definecolor{ablB}{HTML}{ededed}
\definecolor{ablC}{HTML}{ededed}
\definecolor{symBlue}{HTML}{2c6abf}
\newcommand{\chunkmark}{\textcolor{symBlue}{$\boxplus$}}
\newcommand{\kgmark}{\textcolor{symBlue}{$\blacktriangledown$}}
\newcommand{\hgmark}{\textcolor{symBlue}{$\square$}}
\newcommand{\asmark}{\textcolor{symBlue}{$\odot$}}
\newcommand{\pmmark}{\textcolor{symBlue}{$\diamond$}}

\newcommand{\cmark}{\textcolor{updelta}{\ding{51}}}
\newcommand{\xmark}{\textcolor{downdelta}{\ding{55}}}
\newcommand{\best}[1]{\cellcolor{bestcell}\textbf{\underline{#1}}}
\newcommand{\second}[1]{\cellcolor{secondcell}\underline{#1}}
\newcommand{\up}[1]{{\scriptsize\textcolor{updelta}{$\uparrow$#1}}}
\newcommand{\dn}[1]{{\scriptsize\textcolor{downdelta}{$\downarrow$#1}}}
\definecolor{idrFrame}{HTML}{2C5282}
\definecolor{idrBg}{HTML}{EFF4FA}
\definecolor{agentFrame}{HTML}{2F6A4A}
\definecolor{agentBg}{HTML}{F0F6F2}
\definecolor{judgeFrame}{HTML}{6B4C8A}
\definecolor{judgeBg}{HTML}{F4EEF7}

\lstdefinestyle{promptlst}{
  basicstyle=\small\ttfamily,
  breaklines=true,
  breakatwhitespace=true,
  columns=fullflexible,
  keepspaces=true,
  showstringspaces=false,
  upquote=true,
  literate={`}{{\textasciigrave}}1,
  aboveskip=2pt,
  belowskip=2pt,
}

\newtcolorbox{idrpromptbox}[1]{
  enhanced, breakable,
  colback=idrBg, colframe=idrFrame,
  fonttitle=\bfseries\small, coltitle=white,
  title={#1},
  arc=2pt, boxrule=0.8pt,
  left=6pt, right=6pt, top=4pt, bottom=4pt,
  before skip=6pt, after skip=6pt
}
\newtcolorbox{agentpromptbox}[1]{
  enhanced, breakable,
  colback=agentBg, colframe=agentFrame,
  fonttitle=\bfseries\small, coltitle=white,
  title={#1},
  arc=2pt, boxrule=0.8pt,
  left=6pt, right=6pt, top=4pt, bottom=4pt,
  before skip=6pt, after skip=6pt
}
\newtcolorbox{judgepromptbox}[1]{
  enhanced, breakable,
  colback=judgeBg, colframe=judgeFrame,
  fonttitle=\bfseries\small, coltitle=white,
  title={#1},
  arc=2pt, boxrule=0.8pt,
  left=6pt, right=6pt, top=4pt, bottom=4pt,
  before skip=6pt, after skip=6pt
}

\definecolor{designFrame}{HTML}{475569}
\definecolor{designBg}{HTML}{F4F6FB}
\newtcolorbox{designbox}[1]{
  enhanced, breakable,
  colback=designBg, colframe=designFrame,
  fonttitle=\bfseries\small, coltitle=white,
  title={#1},
  arc=2pt, boxrule=0.8pt,
  left=6pt, right=6pt, top=4pt, bottom=4pt,
  before skip=6pt, after skip=6pt
}

\theoremstyle{definition}
\newtheorem{theorem}{Theorem}

\newcommand{\policy}{\pi_\theta}
\newcommand{\oracle}{\mathcal{O}}
\newcommand{\graph}{\mathcal{G}}
\newcommand{\query}{q}
\newcommand{\answer}{a}
\newcommand{\memory}{\mathcal{M}}
\newcommand{\IDR}{\textsc{IDR}}
\newcommand{\PAR}{\textsc{PAR}}
\newcommand{\GPM}{\textsc{GPM}}
\newcommand{\type}{\mathrm{type}}
\newcommand{\sig}{\sigma}
\newcommand{\method}{PatternBloom}

\title{PatternBloom: Empowering Agentic RAG with Externalized RL-Distilled Graph Patterns}

\author{%
  Yi Ge, Heng Huang \\
  Department of Computer Science, 
  University of Maryland, College Park \\
  \texttt{\{yige, heng\}@umd.edu} \\
}

\begin{document}

\maketitle

\begin{abstract}
Reinforcement-learning-trained agents for retrieval-augmented generation retrieve content (passages, knowledge-graph fragments, or hyperedges) and supervise training with outcome rewards on the final answer. Across these paradigms the reasoning route through retrieved content never leaves the language model's hidden state, conflating declarative knowledge with the procedural knowledge that recurs across queries of the same logical type; content-only methods consequently plateau as reasoning depth grows. We introduce \method{}, an agentic RAG framework that shifts the unit of retrieval from content (what to read) to logic (how to reason). A first RL stage scores the agent's constructed evidence graph by a frozen oracle's information gain through the Information-Density Reward (\IDR{}), a size-normalized mutual-information surrogate. High-reward trajectories are distilled into a Graph Pattern Memory (\GPM{}) of type-abstracted reasoning skeletons, and a second RL stage uses the Pattern-Augmented Reward (\PAR{}) to couple policy to memory, turning procedural reasoning into a learnable, externalized structural prior. A 7B \method{} outperforms $27$ baselines by $+15.0$ Avg-OOD EM, with the gain compounding with reasoning depth precisely because procedural knowledge, once distilled, is reused rather than re-derived at every query. We release our code at \url{https://github.com/EllenYiGe/PatternBloom}.
\end{abstract}

\input{sections/1_introduction}
\input{sections/2_related_work}
\input{sections/3_methodology}
\input{sections/4_experiments}
\input{sections/6_conclusion}

\begin{ack}
This work was partially supported by NSF IIS~2347592, 2348169, DBI~2405416, CCF~2348306, CNS~2347617, RISE~2536663. The authors declare no competing interests.
\end{ack}

\bibliographystyle{unsrtnat}
\bibliography{references}


\clearpage
\appendix
\section*{Appendix}
\addcontentsline{toc}{part}{Appendix}

\input{sections/7_appendix}


\newpage
\input{checklist.tex}

\end{document}

%% file: sections/1_introduction.tex
\section{Introduction}
\label{sec:intro}

Retrieval-augmented language model agents must compose retrieved
evidence to answer multi-hop questions over large
corpora~\citep{rag_lewis, hotpotqa, twowiki, musique}. Recent agentic
systems trained with reinforcement
learning (Search-R1~\citep{search_r1}, Graph-R1~\citep{graph_r1},
R1-Searcher~\citep{r1_searcher}) push this further by letting an
agent decide when and what to retrieve, optimizing tool-use policies
with outcome rewards~\citep{deepseek_r1, grpo_math}. Across these
architectures, however, supervision touches only the retrieved units
and the final answer string. Rewards score the correctness of the
answer, but the reasoning trajectory between query and answer (which
entities the agent binds, in which order, along which logical path)
is left to a black-box policy: the agent learns to read better
evidence over time, yet does not learn to traverse evidence better.

\begin{figure*}[t]
\centering
\includegraphics[width=0.92\linewidth]{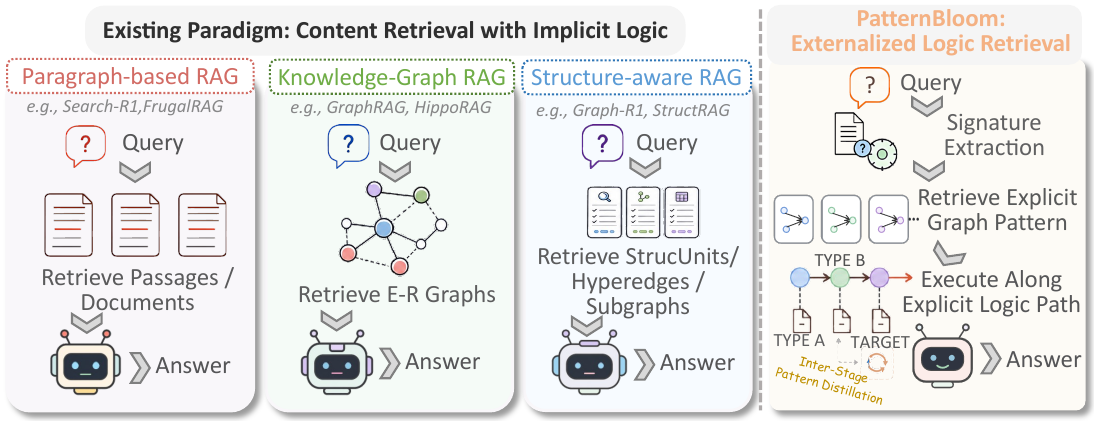}
\caption{\textbf{Content Retrieval paradigm} versus \textbf{Logic Retrieval paradigm}. \emph{Left}: prior paradigms retrieve content units (passages, knowledge-graph fragments, or hyperedges), leaving the reasoning route implicit in policy weights. \emph{Right}: \method{} retrieves a typed reasoning skeleton from an externalized Graph Pattern Memory and traverses it as an explicit logic path.}
\label{fig:teaser}
\end{figure*}

Existing RAG paradigms differ primarily in \emph{what is retrieved}: paragraph-based methods (Search-R1, FrugalRAG~\citep{frugalrag2026}, R1-Searcher) return chunk-level passages; knowledge-graph methods (GraphRAG~\citep{graphrag_ms}, HippoRAG~\citep{hipporag, hipporag2}, LinearRAG~\citep{linearrag2026}, LightRAG~\citep{lightrag}, GFM-RAG~\citep{gfm_rag}) build entity-relation graphs offline and return subgraphs; and structure-aware methods (StructRAG~\citep{structrag}, Graph-R1, Structure-R1~\citep{structure_r1}) retrieve task-adaptive units such as hyperedges or summarized clusters. Despite these different retrieval units, the three paradigms share a common limitation: the reasoning route through the retrieved evidence never leaves the language model's hidden state. Paragraph methods leave entity binding and relation inference to opaque attention; knowledge-graph and structure-aware methods expose entities as graph nodes, but the path through those nodes that solves a given query is still inferred at run time. Figure~\ref{fig:teaser} (left) frames this collectively as \emph{content retrieval}: the agent decides \contentP{what to read} (a passage, a subgraph, or a hyperedge), but the reasoning skeleton that links those units is never itself retrieved. The cost compounds on compositionally complex queries, motivating a retrieval target that is the reasoning route itself.

\paragraph{Why richer content does not solve the problem.}
The plateau on deeper queries is not primarily a retrieval-recall failure: knowledge-graph and structure-aware methods often achieve high entity-level coverage. The bottleneck is procedural. Two queries about different films, ``in which country was the director of \textsc{Black Gold} born?'' and ``where was the director of \textsc{Top Gun} from?'', share an identical logical structure yet are solved from scratch each time: their typed reasoning skeleton, from \textsc{film} to \textsc{person} to \textsc{country}, is encoded only in the policy's hidden activations and discarded when the episode ends. Cognitively, this conflates declarative knowledge (the facts to retrieve) with procedural knowledge (the operations that recur across queries of the same logical type)~\citep{squire1992memory}. The two memory systems are functionally distinct in human reasoning and require different storage and retrieval mechanisms; a retrieval system that augments only declarative memory will, by construction, fail to compound experience across queries that share the same procedural skeleton.

\paragraph{From content retrieval to logic retrieval.} To this end, we introduce \method{}, an agentic RAG system that shifts the retrieval target from content (\contentP{what to read}) to logic (\logicP{how to reason}). The unit of retrieval is a \emph{graph pattern} (Fig.~\ref{fig:teaser}, right): a small directed skeleton over typed entity slots that prescribes how to traverse retrieved evidence rather than what evidence to read. The skeleton is type-abstracted, so the same pattern serves any query of the same logical structure (concrete examples in \S\ref{sec:gpm}), making procedural knowledge an explicit, retrievable object rather than a behavior latent in policy weights. A two-stage RL pipeline trains and exploits this externalized memory. Stage~I uses an Information-Density Reward to supervise the agent on the information content of the constructed evidence graph rather than on the answer string, promoting the graph to a directly supervised quantity. An inter-stage distillation extracts type-abstracted skeletons from high-reward trajectories into a Graph Pattern Memory. Stage~II uses a Pattern-Augmented Reward to couple the policy to the frozen memory, crediting trajectories that traverse retrieved patterns and turning procedural reasoning from a soft prompt hint into a gradient signal. Where prior systems retrieve passages (paragraph methods), subgraphs (knowledge-graph methods), or hyperedges (structure-aware methods), \method{} retrieves a typed reasoning skeleton: the logic itself.

Realizing this paradigm shift surfaces three concrete challenges: (C1) supervising the constructed evidence graph as an information object rather than scoring only the answer string; (C2) externalizing reasoning patterns in a form that is reusable across queries and backbones; and (C3) coupling a frozen pattern store back to the policy as a process-level gradient signal. Our main contributions, summarized below, address (C1)--(C3) respectively via the Information-Density Reward, the Graph Pattern Memory, and the Pattern-Augmented Reward.
\begin{itemize}[leftmargin=12pt, itemsep=2pt]
  \item \textbf{A new retrieval paradigm for agentic RAG: from content to logic.} \method{} reframes agentic RAG by retrieving typed reasoning skeletons rather than content units. Stored in an externalized Graph Pattern Memory (\GPM{}) auto-distilled from a Stage-I RL agent's own high-reward trajectories, these skeletons make procedural knowledge a first-class retrievable artefact: inspectable, frozen at deployment, and reusable across queries and backbones.
  \item \textbf{An information-theoretic two-stage RL framework.} Stage~I supervises the constructed evidence graph itself via the Information-Density Reward (\IDR{}); Theorem~\ref{thm:idr} establishes \IDR{} as a bounded monotone surrogate for size-normalized conditional mutual information, the first information-theoretic reward design for agentic graph construction. Stage~II couples the policy to the frozen \GPM{} via the Pattern-Augmented Reward (\PAR{}), promoting pattern coverage from a soft prompt hint to a process-level gradient signal.
  \item \textbf{Procedural-knowledge accumulation as a new scaling axis.} A 7B \method{} outperforms $27$ baselines by $+15.0$ Avg-OOD EM; the gain compounds with reasoning depth where content-only retrieval saturates, and the same frozen pattern memory transfers across backbone families ($1.5$B--$32$B), establishing pattern retrieval as a scaling axis distinct from larger backbones, deeper retrieval, or richer prompts.
\end{itemize}

%% file: sections/2_related_work.tex
\section{Related Work}
\label{sec:related}

\paragraph{Structured retrieval-augmented generation.}
Retrieval-augmented generation~\citep{rag_lewis,dpr,colbert,fid,bm25} has progressively moved toward structured evidence. Knowledge-graph methods build offline graphs and return subgraphs (GraphRAG~\citep{graphrag_ms}, HippoRAG~\citep{hipporag,hipporag2}, LightRAG~\citep{lightrag}, LinearRAG~\citep{linearrag2026}, GFM-RAG~\citep{gfm_rag}); hypergraph variants extend to $n$-ary facts (HyperGraphRAG~\citep{hypergraphrag}, Graph-R1~\citep{graph_r1}); adaptive-structure methods select per-query formats (StructRAG~\citep{structrag}, Structure-R1~\citep{structure_r1}). Across this line the structure is fixed at retrieval time, and the agent receives content fragments rather than a reasoning path.

\paragraph{Reinforcement learning for agentic RAG.}
A complementary line trains agents end-to-end with policy gradients~\citep{ppo,dpo,deepseek_r1,grpo_math}: Search-R1~\citep{search_r1}, R1-Searcher~\citep{r1_searcher}, FrugalRAG~\citep{frugalrag2026}, Graph-R1, Structure-R1, and ReSearch~\citep{chen2025research} share answer-level outcome rewards; prompted variants couple iterative retrieval with chain-of-thought~\citep{ircot,react,selfrag,toolformer,cot}.

Two constraints bottleneck both lines: (i)~supervision is surface-form, scoring the answer string but never the evidence graph as an information object; (ii)~the retrieval unit is content tied to specific entities, not the procedural skeleton that recurs across queries of the same logical type. \method{} addresses both via an information-theoretic graph reward and an externalized Graph Pattern Memory of type-abstracted skeletons; a method-by-method comparison appears in Appendix~\ref{app:related}.

%% file: sections/3_methodology.tex
\section{\method: Logic-Enhanced RL with Externalized Graph Pattern Memory}
\label{sec:approach}

\begin{figure*}[t!]
\centering
\includegraphics[width=\linewidth]{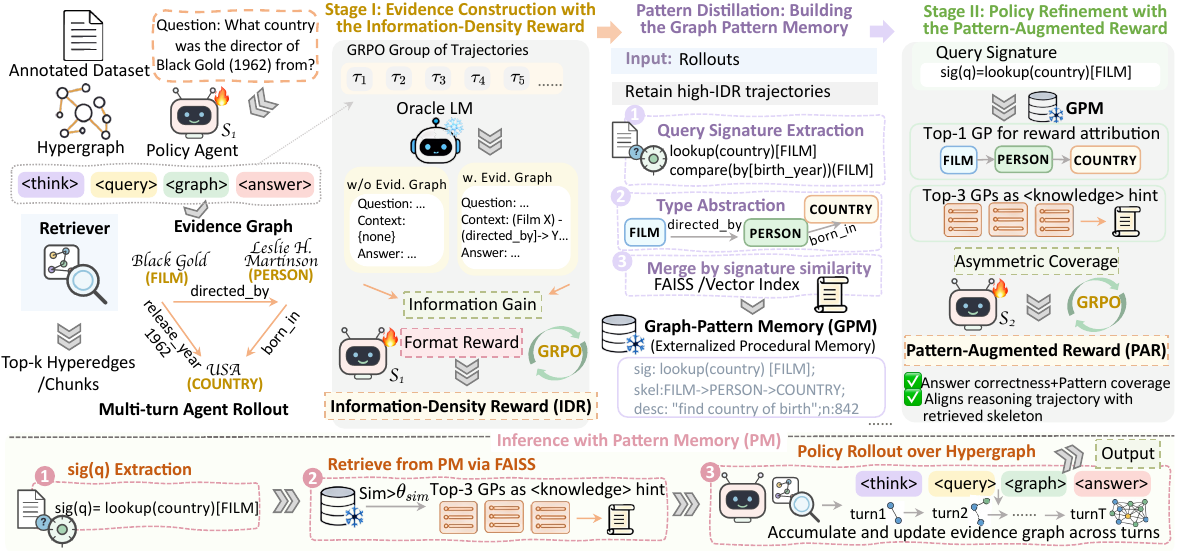}
\caption{\method{} pipeline. \emph{Top:} training with intermediate pattern distillation between Stage~I (IDR) and Stage~II (PAR). \emph{Bottom:} deployment uses only trained policy and frozen pattern memory.}
\label{fig:pipeline}
\end{figure*}

Section~\ref{sec:related} identified two constraints that bottleneck prior agentic RAG: surface-form supervision that never interrogates the constructed evidence graph, and a content-only retrieval target that leaves procedural reasoning latent in policy weights. \method{} is built around one principle that addresses both: supervise the agent on \emph{what} evidence it gathers and \emph{how} it traverses that evidence, with each component an externalized, learnable object rather than a behavior to be hoped for. Stage~I (\S\ref{sec:idr}) targets the first constraint via the Information-Density Reward (\IDR{}), which scores the evidence graph's information content under a frozen oracle and so promotes the graph itself to a directly supervised quantity. Trajectories from the Stage-I best checkpoint are distilled (\S\ref{sec:gpm}) into a Graph Pattern Memory (\GPM{}) of type-abstracted reasoning skeletons; Stage~II (\S\ref{sec:par}) couples the frozen \GPM{} to the policy via the Pattern-Augmented Reward (\PAR{}), turning procedural reasoning into a retrievable, learnable structural prior and addressing the second constraint. Figure~\ref{fig:pipeline} shows the pipeline; Algorithm~\ref{alg:patternbloom} (Appendix~\ref{app:algorithm}) gives the formal procedure.

\subsection{Preliminaries}
\label{sec:problem}

We consider multi-hop QA over a corpus $\mathcal{C}$ of Wikipedia chunks, with dataset $\mathcal{D} = \{(\query_i, \answer^*_i)\}$ where each query $\query$ requires reasoning over two or more retrieved facts. We pre-build a hypergraph index $H = (V, E_{\text{hyper}}, X_{\text{text}})$ over $\mathcal{C}$ following \citet{hipporag, graph_r1}: entity nodes $V$ are connected by hyperedges $E_{\text{hyper}}$ that group facts from each chunk, and chunk text $X_{\text{text}}$ is preserved for retrieval (Appendix~\ref{app:training:index}). $n$-ary hyperedges preserve the intra-chunk co-occurrence that binary triples discard, exposing the typed neighborhoods that pattern abstraction (\S\ref{sec:gpm}) will exploit. The agent policy $\policy$ is a tool-use language model that interleaves $\langle\textsf{think}\rangle$, $\langle\textsf{query}\rangle$, $\langle\textsf{graph}\rangle$, $\langle\textsf{answer}\rangle$ blocks, accumulating an evidence graph $\graph \subseteq H$ over multiple turns. The oracle $\oracle$ (Qwen2.5-1.5B-Instruct~\citep{qwen25}) is held \emph{frozen} and used only at training time to score evidence-graph quality via $\log P_\oracle(\cdot \mid \query, \cdot)$: a frozen scorer keeps the reward landscape stationary across both stages and rules out reward-gaming that an online critic invites. The pattern memory $\memory$ is defined precisely in \S\ref{sec:gpm}.

\paragraph{Per-stage GRPO objective.}
\method{}'s two rewards target complementary failure modes: $R_{\IDR}$ (Stage~I) addresses what the policy fails to gather, and $R_{\PAR}$ (Stage~II) addresses how the policy fails to structure its reasoning. We instantiate both stages with GRPO~\citep{grpo_math}---group-relative advantages remove the value-function critic, which becomes high-variance under the sparse trajectory-level rewards typical of long tool-use rollouts---through a single policy-gradient template with $R$ supplying the per-stage signal:
\begin{equation}\label{eq:grpo_obj}
\begin{split}
\mathcal{L}_{\text{\method}}(\theta;\, R) = {}& \mathbb{E}_{\query,\{\tau_i\}}\!\Bigl[\tfrac{1}{G}\textstyle\sum_i \min\bigl(r_i(\theta)\,\hat{A}_i,\, \mathrm{clip}(r_i(\theta), 1{-}\epsilon, 1{+}\epsilon)\,\hat{A}_i\bigr)\Bigr] \\
& - \beta\, D_{\mathrm{KL}}\bigl[\policy \,\|\, \pi_{\mathrm{ref}}\bigr],
\end{split}
\end{equation}
where $\hat{A}_i = \bigl(R(\tau_i) - \mathrm{mean}_G R\bigr)/\bigl(\mathrm{std}_G R + \delta\bigr)$ is the within-group standardized advantage, $r_i(\theta)$ is the per-rollout importance ratio, and $\pi_{\mathrm{ref}}$ is reset between stages (the SFT backbone for Stage~I, the Stage-I best policy for Stage~II) so each stage is anchored to the previous stage's gains. Both rewards are bounded in $[-1, 1]$ via a binary format gate. Stage~I sets $R = R_{\IDR}$ (Eq.~\ref{eq:idr}); Stage~II sets $R = R_{\PAR}$ (Eq.~\ref{eq:par}); hyperparameters are listed in Appendix~\ref{app:training:hp}.

\subsection{Stage I: Evidence Construction with the Information-Density Reward}
\label{sec:idr}

Existing RL-RAG agents supervise training with answer-level signals on the predicted string~\citep{search_r1, r1_searcher} or entity-presence checks on the constructed graph~\citep{graph_r1}. None interrogate the graph's information content as a whole: a $20$-triple noisy graph and a $3$-triple clean one receive the same reward provided both yield a matching answer. This blindness leaves $\policy$ with no signal to prefer minimum-sufficient evidence over verbose-but-correct rollouts, and the gap widens with reasoning depth (Appendix~\ref{app:hopdepth}). \IDR{} closes this gap by promoting the constructed graph itself to the supervised quantity.

Let $\mathrm{ser}(\graph)$ serialize the evidence graph as a sequence of typed triples. The \emph{information gain} of $\graph$ relative to the empty graph is
\begin{equation}\label{eq:dlogp}
\Delta\log p(\graph) \coloneqq
\log P_{\oracle}\bigl(\answer^* \mid \query, \mathrm{ser}(\graph)\bigr)
- \log P_{\oracle}\bigl(\answer^* \mid \query, \emptyset\bigr),
\end{equation}
which equals the conditional pointwise mutual information $\mathrm{PMI}(\graph; \answer^* \mid \query)$ between graph and answer given the query (cf.~Eq.~\ref{eq:pmi}). The Information-Density Reward wraps the per-triple-normalized gain in a sigmoid, gates by the format reward $R_{\text{fmt}}$ so malformed trajectories collapse to $-1$, and rewards trajectories that achieve high information density:
\begin{equation}\label{eq:idr}
R_{\IDR}(\tau) = -1 + R_{\text{fmt}}(\tau) + \mathds{1}\{R_{\text{fmt}}=1\} \cdot
\sigma\!\left( \frac{\Delta\log p(\graph)}{\max(|\graph|, 1) \cdot T} \right),
\end{equation}
with temperature $T=1$. Three design choices in Eq.~\ref{eq:idr} are deliberate. \emph{(i)~Size normalization.} Dividing by $|\graph|$ favors short anchor-to-answer paths: low-information triples grow $|\graph|$ faster than $\Delta\log p$, so verbose-but-correct rollouts strictly lose to minimum-sufficient ones, closing a loophole that entity-presence rewards leave open~\citep{graph_r1}. \emph{(ii)~Sigmoid wrapping.} The sigmoid bounds the reward in $[0, 1]$ so within-group GRPO advantages remain well-scaled across queries whose absolute $\Delta\log p$ scale varies by orders of magnitude. \emph{(iii)~Format gating.} The $R_{\text{fmt}}$ multiplier collapses malformed trajectories to $-1$, so any density improvement is conditional on a well-formed graph and the policy cannot trade structural validity for content. Theorem~\ref{thm:idr} (Appendix~\ref{app:proofs}) shows that maximizing $R_{\IDR}$ via GRPO is consistent with maximizing the size-normalized conditional mutual information $I(\graph; \answer^* \mid \query)/(|\graph|\cdot T)$, with the divisive normalizer realizing an adaptive Information-Bottleneck multiplier~\citep{tishby_ib,tishby2015deep} $\beta_{\text{eff}}=1/(|\graph|\cdot T)$. \IDR{} thereby provides the first information-theoretic reward design for agentic graph construction, replacing the surface-form heuristics that bottleneck prior work with a principled signal that scores the graph as a whole.

\subsection{Pattern Distillation: Building the Externalized Procedural Memory}
\label{sec:gpm}

A retrieved triple like $(\textsc{Black Gold}, \texttt{directed\_by}, \textsc{Leslie Martinson})$ is query-specific and never recurs, but many queries share the same logical structure: ``given a film, find the director's nationality'' applies across thousands of film--director--country instances. The procedural component of multi-hop reasoning---\emph{how} types of facts compose---recurs while the instances change, yet prior systems leave this regularity implicit in policy weights (\S\ref{sec:related}). \method{} lifts each instance triple to a type-abstracted skeleton $(\textsc{film}, \texttt{directed\_by}, \textsc{person}), (\textsc{person}, \texttt{born\_in}, \textsc{country})$ that is reusable across all instances of the same query type, and stores these skeletons in an external memory auto-distilled from the policy's own high-reward Stage-I trajectories. Three properties follow from this construction: \emph{type abstraction} makes each entry an equivalence class over instances and so amortizes the cost of one query's reasoning across thousands of structurally similar ones; \emph{externalization} makes the memory inspectable, freezable, and reusable across backbones; and \emph{self-distillation} keeps the memory aligned with what this policy actually solves, rather than a hand-curated wishlist.

\paragraph{Graph pattern.}
A graph pattern $P$ is a tuple $(\sig, \mathrm{skel}, \mathrm{desc}, n)$ where $\sig \in \mathcal{S}$ is a query signature, $\mathrm{skel}$ is a connected directed graph of typed entity nodes (the reasoning skeleton), $\mathrm{desc}$ is a short natural-language descriptor, and $n$ is a usage count. The Graph Pattern Memory $\memory$ is the learned external store
\begin{equation}\label{eq:gpm}
\memory = \bigl\{\bigl(\sig_i,\, \mathrm{skel}_i,\, \mathrm{desc}_i,\, n_i\bigr)\bigr\}_{i=1}^{|\memory|},
\end{equation}
populated by distilling Stage-I trajectories. Signatures are extracted by a deterministic parser (e.g., \texttt{lookup(country)[FILM]}); entity types in the skeleton come from the type-abstraction step, and descriptors are auto-generated from the signature and a representative answer instance.

\paragraph{Coverage and retrieval.}
At Stage~II, patterns are retrieved by encoding $\sig(\query)$ with a sentence encoder and ranking $\memory$ by cosine similarity in a FAISS~\citep{faiss} index. The trajectory--pattern coverage
\begin{equation}\label{eq:cpattern}
\mathcal{C}(\tau, P) = \frac{|\,\mathrm{abs}(\tau) \cap P.\mathrm{skel}\,|}{|P.\mathrm{skel}|},
\quad \mathrm{abs}(\tau) \coloneqq \{(\type(h), r, \type(t)) : (h, r, t) \in \tau\},
\end{equation}
divides by the skeleton size rather than by the union, so a trajectory that contains the required reasoning steps plus exploratory extras receives full credit. We use the top-$1$ match $P^*_\query \in \memory$ as the coverage signal in $R_{\PAR}$ and feed the top-$3$ retrieved patterns to the agent's prompt as a $\langle\textsf{knowledge}\rangle$ hint, allowing the policy to compare alternative skeletons before committing to a trajectory.

\paragraph{Memory construction.}
The memory $\memory_0$ is built between Stage~I and Stage~II from high-IDR trajectories ($R_{\IDR} \geq 0.85$, the empirical top-quartile threshold; Appendix~\ref{app:gpm}); two patterns merge when their signature embeddings have cosine similarity $\geq \theta_{\mathrm{merge}}=0.85$. We freeze $\memory_0$ throughout Stage~II and at inference: a non-stationary memory chasing a non-stationary policy invites collapse, while a frozen memory yields a stationary reward signal during refinement and a deterministic retrieval store at deployment. The same frozen $27$-skeleton memory (Tab.~\ref{tab:gpm_stats}) then transfers across distribution shift, a test query walks a skeleton distilled purely from training (Fig.~\ref{fig:case_study_main}), evidencing that procedural knowledge, once externalized, generalizes beyond its training distribution.

\subsection{Stage II: Policy Refinement with the Pattern-Augmented Reward}
\label{sec:par}

Feeding retrieved patterns to the policy's prompt provides only a soft prior, with no gradient signal pushing the trajectory to align with the retrieved skeleton: the policy is free to ignore the hint and revert to whatever idiosyncratic chain it has already learned, especially when answer-level reward is uninformative. \method{} promotes pattern coverage from a soft prior to a process-level reward by combining answer-level $F_1$ with the trajectory--pattern coverage of Eq.~\ref{eq:cpattern}:
\begin{equation}\label{eq:par}
R_{\PAR}(\tau) = -1 + R_{\text{fmt}}(\tau) + \mathds{1}\{R_{\text{fmt}}=1\}
\cdot \bigl[\, (1-\lambda_P)\, F_1(\hat{\answer}, \answer^*) + \lambda_P\, \mathcal{C}(\tau, P^*_{\query})\,\bigr].
\end{equation}
The convex combination ensures pattern coverage acts as a structural prior that shapes \emph{how} the policy reasons without overriding the answer signal that defines task success. The shaping weight $\lambda_P \in [0, 0.3]$ keeps answer correctness dominant, in line with the bounded-potential shaping principle of~\citet{ng1999reward}, which guarantees policy invariance under shaping rewards small relative to the task reward. We select $\lambda_P = 0.2$ as the default through a sweep over $\{0, 0.1, 0.2, 0.5\}$ on a held-out 1k-query slice (Appendix~\ref{app:par}); values above $0.3$ destabilize the IDR-learned graph-construction behavior because pattern coverage begins to dominate the answer-level signal, and we therefore impose this empirical bound. Setting $\lambda_P = 0$ recovers a pure $F_1$ reward. \method{} warm-starts $\policy$ from $\pi_{\theta^*_1}$ and continues maximizing $\mathcal{L}_{\text{\method}}(\theta; R_{\PAR})$ (Eq.~\ref{eq:grpo_obj}); at deployment, $\pi_{\theta^*_2}$ and the frozen $\memory_0$ operate over the same hypergraph $H$ (Appendix~\ref{app:inference_pipeline}). Together, \IDR{} and \PAR{} realize the paradigm shift of \S\ref{sec:intro}: both the constructed evidence graph and the retrieved reasoning skeleton become supervisable, externalized objects rather than behaviors latent in policy weights, and the gain compounds with reasoning depth precisely because procedural knowledge, once distilled, is reused rather than re-derived at every query.

%% file: sections/4_experiments.tex
\section{Experiments}
\label{sec:exp}

\subsection{Experimental Setup}
\label{sec:exp_setup}

\paragraph{Datasets.}
Seven QA benchmarks split as in-distribution (HotpotQA~\citep{hotpotqa}, 2WikiMultiHopQA~\citep{twowiki}; $14$K FlashRAG training queries, seed $42$) and out-of-distribution (MuSiQue~\citep{musique}, NQ~\citep{kwiatkowski2019nq}, TriviaQA~\citep{triviaqa}, PopQA~\citep{popqa}, Bamboogle~\citep{bamboogle_self_ask}). We sample $1{,}000$ test instances per dataset (seed $40$); Bamboogle uses its full $125$ items. The $5$-of-$7$ OOD majority is strictly harder than the IID-only protocol of Graph-R1~\citep{graph_r1}.

\paragraph{Baselines.}
$27$ baselines span four families: training-free RAG on GPT-4o-mini (HippoRAG~2~\citep{hipporag2}, GraphRAG~\citep{graphrag_ms}, LightRAG~\citep{lightrag}, StructRAG~\citep{structrag}, LinearRAG~\citep{linearrag2026}, GFM-RAG~\citep{gfm_rag}, HyperGraphRAG~\citep{hypergraphrag}); RL-trained chunk retrieval (Search-R1~\citep{search_r1}, R1-Searcher~\citep{r1_searcher}, FrugalRAG~\citep{frugalrag2026}); RL-trained graph retrieval (Graph-R1~\citep{graph_r1}, Structure-R1~\citep{structure_r1}); plus prompt-engineered ($72$B) and large-scale ($14$B--$\geq30$B) reference panels (Tab.~\ref{tab:main}). RL methods share identical training data and seeds; \method{} occupies a previously unfilled cell: a trained agent over a graph substrate with procedural memory.

\paragraph{Metrics, backbones, compute.}
We report answer correctness (EM, F1, Contain-Acc), retrieval quality (R@5, Evidence Recall), pattern faithfulness (RPF, P-Cov; \S\ref{sec:exp_reasoning}), efficiency, and human + LLM-judge scores. Backbones: Qwen2.5-7B-Instruct~\citep{qwen25} (primary), Qwen3-8B-Instruct~\citep{qwen3} (cross-backbone), four Qwen2.5 sizes ($1.5$B/$7$B/$14$B/$32$B) for scaling. Full settings and metric definitions: Appendix~\ref{app:training:hp},~\ref{tab:metric_panel}.

\subsection{Main Results}
\label{sec:exp_main}

\input{table/4_exp/tab1_main}

\method{} (Q2.5-7B) is rank-$1$ on EM and F1 across all seven datasets in Tab.~\ref{tab:main}, leading R1-Searcher by $15.0$ Avg-OOD EM and the $72$B prompt-engineered RAG panel by $17.1$ Avg-OOD EM at one-tenth the parameter count. The same frozen \GPM{} transfers to a Qwen3-8B backbone, where \method{} leads every $8$B entry on Avg-OOD by $11.3$ EM, indicating the gain originates in the procedural-memory mechanism rather than backbone-specific tuning. The IID-to-OOD gap stays under $1.5$ EM on either backbone (against $4.1$ and $3.3$ EM for Search-R1 and Graph-R1), and \method{} is rank-$1$ on every Avg-OOD metric (EM, F1, Contain-Acc, R@5, Evidence Recall), so the gain transfers across the five out-of-distribution test sets without overfitting to the IID training corpus.

\subsection{Component Analysis}
\label{sec:exp_ablation}

\input{table/4_exp/tab2_ablation}

The three components are jointly necessary. Removing \PAR{} drops F1 from $55.9$ to $44.1$, a larger loss than disabling \GPM{} ($46.6$) or skipping \IDR{} ($52.9$): \PAR{} is the only mechanism supplying a process-level gradient signal, \GPM{} the structural prior it exploits, and \IDR{}'s standalone effect is partially recovered by Stage~II. The pattern-coverage weight is concave around $\lambda_P=0.2$ ($55.9$ F1; $\lambda_P\in\{0,0.1,0.5\}$ each lose $10$--$13$ F1), confirming the bound $\lambda_P\leq0.3$ derived in \S\ref{sec:par} from \citet{ng1999reward}. Memory access is all-or-nothing: top-$1$/$5$/$10$ retrieval cluster around $43$ F1 while the full $|\memory|=27$ pool jumps to $55.9$ F1, so the gain comes from the pool's joint coverage rather than any single dominant skeleton.

\subsection{Reasoning Quality}
\label{sec:exp_reasoning}
\input{table/4_exp/tab3_pattern_quality}

We complement answer-level scores with two pattern-quality signals introduced in this work: \emph{RPF} (Reasoning Path Faithfulness, fraction of trajectory turns invoking $\geq 1$ stored pattern) and \emph{P-Cov} (Pattern Coverage, fraction of trajectories whose retrieved pattern matches the oracle-best signature).
Both drop monotonically as \method{} components are removed and reach zero with GPM disabled (by construction); P-Cov correlates with answer EM ($\rho=0.87$ across the four \method{} variants), evidence that pattern coverage is a faithful surrogate for reasoning quality rather than a separate dimension.

\begin{figure}[htbp]
\centering
\begin{subfigure}[b]{0.290\linewidth}
  \centering
  \includegraphics[width=\linewidth]{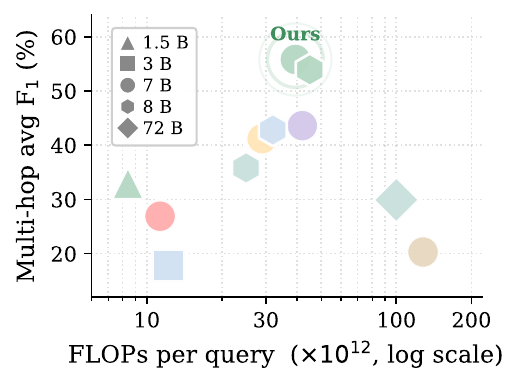}
  \caption{Cost-quality analysis.}
  \label{fig:perf_profile_cost}
\end{subfigure}\hfill
\begin{subfigure}[b]{0.275\linewidth}
  \centering
  \vspace*{8pt}
  \includegraphics[width=\linewidth]{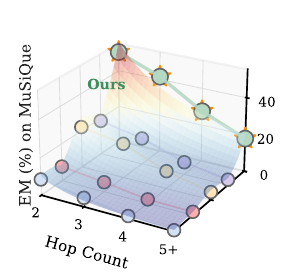}
  \caption{Reasoning-depth scaling.}
  \label{fig:perf_profile_depth}
\end{subfigure}\hfill
\begin{subfigure}[b]{0.380\linewidth}
  \centering
  \includegraphics[width=\linewidth]{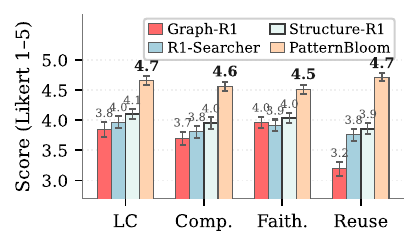}
  \caption{Human evaluation.}
  \label{fig:human_study}
\end{subfigure}
\caption{Performance and reasoning quality across three views: (a) cost-quality trade-off, (b) reasoning-depth scaling on MuSiQue, (c) human evaluation.}
\label{fig:perf_profile}
\end{figure}

\subsection{Multi-Size Scaling and Cross-Backbone Transfer}
\label{sec:exp_scaling}

\method{} retrains at four Qwen2.5 sizes ($1.5$B/$7$B/$14$B/$32$B) and one Qwen3-8B variant (per-size scores in Tab.~\ref{tab:main} panels (b)--(e); discussion in Appendix~\ref{app:scaling}) and ranks first on the average column at every size band. The $1.5$B variant attains $28.4$ Avg.\ EM, competitive with $7$B chunk-RL baselines (Search-R1 $7$B at $21.6$) at one-quarter the parameter count, while \method{} (Q3-8B) leads all $8$B entries on Avg-OOD ($43.4$ EM), confirming cross-backbone transfer. Inference latency stays within $0.4$\,s of Graph-R1 (Appendix~\ref{app:cost}). Fig.~\ref{fig:perf_profile} consolidates the Pareto: \method{} sits on the cost-quality frontier (a) and attains the flattest depth-decay ($\alpha=-1.10$ vs.\ Search-R1's $-2.91$, b), substantiating the architectural claim of \S\ref{sec:intro}: procedural memory is an additive architectural component, not a backbone-specific tuning trick.

\subsection{Behavior, Failures, and Human Study}
\label{sec:exp_analysis}

\input{table/4_exp/tab_turns_main}

\paragraph{Question-type behavior.}
Against Graph-R1 (the only directly comparable RL-trained graph-RAG agent), \method{} uses $0.4$--$0.9$ fewer turns per query in every category (Tab.~\ref{tab:turns_per_category}): typed pattern templates replace exploratory iterations with one-shot lookups. Per-category EM/F1 and hop-stratified MuSiQue: Appendix~\ref{app:category},~\ref{app:hopdepth}.

\paragraph{Residual error categories.}
Errors group into four categories (Appendix~\ref{app:failure}): three generic to graph-RAG (entity disambiguation, numerical reasoning, predicate canonicalisation), and one specific to procedural memory (signature mismatch), addressed by an inference-time pattern-admission extension (\S\ref{app:limitations}).

\begin{figure}[htbp]
\centering
\includegraphics[width=\linewidth]{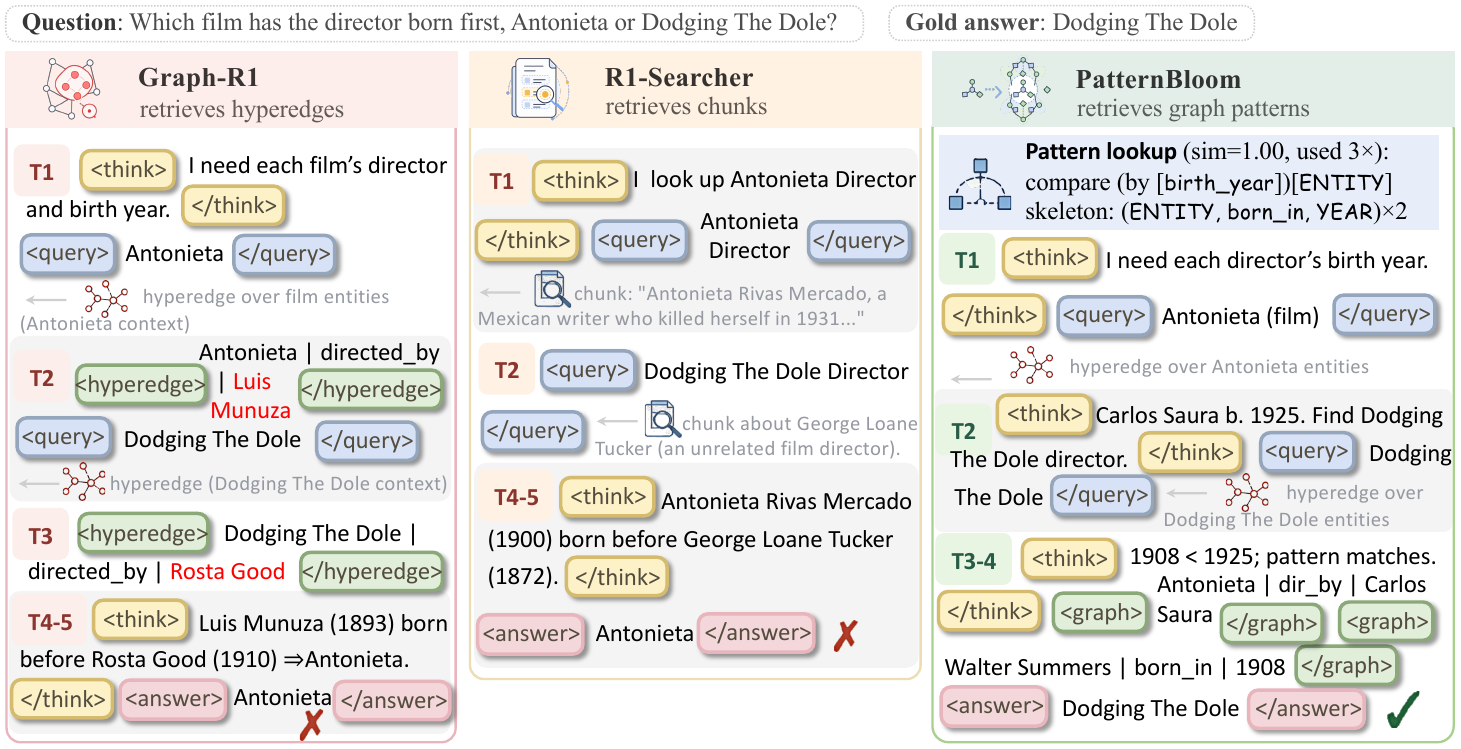}
\caption{Case study on a 4-hop query. Structure-based retrieval fabricates both directors' names, and chunk-based retrieval confuses the film Antonieta with the writer Antonieta Rivas Mercado, while \method{}'s typed skeleton fixes what to look up for each film.}
\label{fig:case_study_main}
\end{figure}

\paragraph{Human study and case illustrations.}
Fig.~\ref{fig:human_study} reports a blind human study on $30$ random queries: \method{} leads on every $1$--$5$ Likert criterion, with the largest gain $+0.85$ on Pattern Reusability over Structure-R1 and the highest inter-annotator agreement ($\alpha=0.74$). Fig.~\ref{fig:case_study_main} contrasts the three paradigms qualitatively: structure-based retrieval fabricates names, chunk-based retrieval substitutes the query's subject, and \method{}'s typed-skeleton retrieval anchors the trajectory to retrieved structure. Trajectories and additional cases: Appendix~\ref{app:cases}.

%% file: table/4_exp/tab1_main.tex
\begin{table*}[t]
\centering
\caption{Main results on seven QA benchmarks, organized as four \emph{Multi-Hop} datasets (HotpotQA, 2WikiMHQA, MuSiQue, Bamboogle) and three \emph{Single-Hop} factoid datasets (NQ, TriviaQA, PopQA). \textsc{IID} marks the two in-distribution training datasets; \textsc{OOD} marks the five held-out datasets. Retrieval representation: \chunkmark\ chunk, \kgmark\ knowledge graph, \hgmark\ hypergraph, \asmark\ adaptive structure, \pmmark\ procedural memory. $^{\ddagger}$ marks \textbf{reported} data; remaining rows are our \textbf{reproduction}s.}
\label{tab:main}
\setlength{\tabcolsep}{2pt}
\renewcommand{\arraystretch}{0.95}
\resizebox{\linewidth}{!}{%
\begin{tabular}{l|cccccccc|cccccc|ccc|ccc}
\toprule
& \multicolumn{8}{c|}{\textbf{Multi-Hop QA}}
& \multicolumn{6}{c|}{\textbf{Single-Hop QA}}
& \multicolumn{3}{c|}{\textbf{Avg-IID}}
& \multicolumn{3}{c}{\textbf{Avg-OOD}} \\
\cmidrule(lr){2-9} \cmidrule(lr){10-15} \cmidrule(lr){16-18} \cmidrule(lr){19-21}
& \multicolumn{2}{c}{\textbf{2Wiki}\textsuperscript{\textsc{IID}}}
& \multicolumn{2}{c}{\textbf{HotpotQA}\textsuperscript{\textsc{IID}}}
& \multicolumn{2}{c}{\textbf{MuSiQue}\textsuperscript{\textsc{OOD}}}
& \multicolumn{2}{c|}{\textbf{Bamboogle}\textsuperscript{\textsc{OOD}}}
& \multicolumn{2}{c}{\textbf{NQ}\textsuperscript{\textsc{OOD}}}
& \multicolumn{2}{c}{\textbf{TriviaQA}\textsuperscript{\textsc{OOD}}}
& \multicolumn{2}{c|}{\textbf{PopQA}\textsuperscript{\textsc{OOD}}}
& & & & & & \\
\textbf{Method} & EM & F1 & EM & F1 & EM & F1 & EM & F1 & EM & F1 & EM & F1 & EM & F1
& EM & F1 & CA
& EM & F1 & CA \\
\midrule
\rowcolor{majGray}\multicolumn{21}{c}{\textsc{\textbf{Vanilla inference-only Baselines (w/ Qwen2.5-7B-Instruct)}}} \\
\midrule
Direct {\scriptsize(no retrieval)}               & 21.7 & 27.7 & 14.2 & 22.8 &  2.8 &  9.4 &  4.0 & 14.4 &  4.6 & 13.7 & 28.0 & 38.5 &  9.2 & 14.4 & 18.0 & 25.3 & 25.8 &  9.7 & 18.1 & 22.2 \\
\chunkmark\ Naive RAG {\scriptsize(E5 chunk top-5)} & 29.7 & 39.5 & 24.3 & 34.9 &  3.7 & 10.5 & 12.0 & 20.2 & 10.9 & 22.7 & 43.8 & 55.3 & 28.0 & 37.4 & 27.0 & 37.2 & 44.0 & 19.7 & 29.2 & 41.6 \\
\midrule
\rowcolor{majGray}\multicolumn{21}{c}{\textsc{\textbf{Prompt-Engineered RAG}}} \\
\midrule
\rowcolor{subA}\multicolumn{21}{l}{\textit{(a) Proprietary closed-source generator (w/ GPT-4o-mini)}} \\
\midrule
\kgmark\ HippoRAG~2          & 18.6 & 22.0 & 24.0 & 33.9 & 27.5 & 38.2 & 39.2 & 48.9 & 26.0 & 37.1 & 50.3 & 59.1 & 14.3 & 17.2 & 21.3 & 27.9 & 27.3 & 31.5 & 40.1 & 39.1 \\
\asmark\ StructRAG           &  9.8 & 13.5 & 29.1 & 39.6 & 11.2 & 18.5 & 39.2 & 49.2 & 36.3 & 51.3 & 66.0 & 74.3 & 40.7 & 49.7 & 19.5 & 26.6 & 24.1 & 38.7 & 48.6 & 47.5 \\
\kgmark\ LinearRAG           & 16.3 & 21.0 & 19.4 & 29.6 & 16.0 & 25.5 & 26.4 & 38.1 & 23.5 & 34.1 & 50.8 & 59.3 & 13.4 & 16.8 & 17.9 & 25.3 & 27.4 & 26.0 & 34.8 & 34.6 \\
\kgmark\ GraphRAG            &  8.8 & 10.0 & 19.3 & 27.1 & 11.5 & 18.0 & 37.6 & 48.1 & 26.9 & 38.7 & 51.9 & 58.9 & 16.9 & 19.4 & 14.1 & 18.6 & 17.6 & 29.0 & 36.6 & 33.9 \\
\kgmark\ LightRAG            &  5.2 &  6.2 & 10.5 & 13.6 & 20.7 & 28.0 & 40.0 & 45.7 & 15.7 & 21.9 & 24.2 & 27.6 &  5.0 &  6.8 &  7.8 &  9.9 & 12.0 & 21.1 & 26.0 & 24.4 \\
\kgmark\ GFM-RAG             & 15.3 & 17.9 & 14.3 & 22.1 & 25.9 & 37.1 & 30.4 & 41.7 & 21.1 & 33.5 & 56.2 & 65.8 & 15.1 & 19.3 & 14.8 & 20.0 & 20.6 & 29.7 & 39.5 & 40.3 \\
\hgmark\ HyperGraphRAG       & 19.8 & 27.9 & 13.3 & 23.8 &  7.7 & 22.2 & 24.8 & 37.3 & 11.1 & 25.8 & 41.2 & 54.6 &  8.0 & 16.6 & 16.6 & 25.9 & 30.1 & 18.6 & 31.3 & 41.6 \\
\midrule
\rowcolor{subB}\multicolumn{21}{l}{\textit{(b) Open-sourced XL-size generator (w/ Qwen2.5-72B-Instruct-AWQ)}} \\
\midrule
\kgmark\ HippoRAG~2          & 10.0 & 13.1 & 18.4 & 23.9 & 24.6 & 34.2 & 39.2 & 48.3 & 23.5 & 33.7 & 41.5 & 48.6 & 12.2 & 14.3 & 14.2 & 18.5 & 19.4 & 28.2 & 35.8 & 35.2 \\
\kgmark\ GraphRAG            &  6.2 &  7.5 & 19.8 & 26.5 & 12.6 & 19.0 & 40.8 & 50.4 & 29.5 & 41.1 & 54.8 & 62.2 & 17.3 & 19.8 & 13.0 & 17.0 & 16.2 & 31.0 & 38.5 & 36.1 \\
\asmark\ StructRAG           &  6.6 &  8.1 & 24.3 & 32.4 &  6.7 & 12.6 & 45.6 & 56.5 & 26.3 & 40.2 & 58.9 & 67.7 & 22.0 & 25.6 & 15.4 & 20.2 & 20.2 & 31.9 & 40.5 & 38.7 \\
\midrule
\rowcolor{majGray}\multicolumn{21}{c}{\textsc{\textbf{Trained Agentic RAG}}} \\
\midrule
\rowcolor{subC}\multicolumn{21}{l}{\textit{(c) $<\!10$B parameters}} \\
\midrule
R1 {\scriptsize(7B)}                        & 29.5 & 34.9 & 21.6 & 29.4 &  5.5 & 14.2 & 32.0 & 38.3 & 15.9 & 25.3 & 43.6 & 50.7 & 14.5 & 17.8 & 25.6 & 32.1 & 39.6 & 22.3 & 29.3 & 30.5 \\
\chunkmark\ Search-R1 {\scriptsize(7B)}     & 34.5 & 40.2 & 33.5 & 44.9 & 11.5 & 19.2 & 40.8 & 51.5 & 32.3 & 46.7 & 70.6 & 79.1 & 35.4 & 40.4 & 34.0 & 42.6 & 39.7 & 38.1 & 47.4 & 44.8 \\
\chunkmark\ R1-Searcher {\scriptsize(7B)}   & 35.3 & 41.3 & 33.7 & 44.6 & 15.8 & 24.3 & 44.0 & 54.7 & 29.5 & 38.5 & 48.6 & 56.0 & 31.9 & 36.9 & 34.5 & 43.0 & \best{63.7} & 34.0 & 42.1 & \second{54.5} \\
\chunkmark\ FrugalRAG {\scriptsize(7B)}     & 15.8 & 21.5 & 15.9 & 24.7 &  3.4 &  8.9 & 17.6 & 26.0 & 15.7 & 20.1 & 41.1 & 49.6 & 16.2 & 25.6 & 15.9 & 23.1 & 23.3 & 18.8 & 26.0 & 25.1 \\
\hgmark\ Graph-R1 {\scriptsize(7B)}         & 27.3 & 32.5 & 20.8 & 27.6 &  4.6 & 12.1 & 23.2 & 35.3 & 17.5 & 27.5 & 43.0 & 49.9 & 15.8 & 19.5 & 24.1 & 30.1 & 39.0 & 20.8 & 28.9 & 29.3 \\
\asmark\ Structure-R1 {\scriptsize(7B)} & \best{69.2} & \best{75.9} & \second{38.0} & 48.9 &  9.9 & 17.5 & 24.8 & 32.1 & 32.4 & 43.9 & 61.7 & 69.4 & \second{39.5} & \second{46.5} & \best{53.6} & \best{62.4} & \second{59.0} & 33.7 & 41.9 & 39.6 \\
R1 {\scriptsize(8B)}                        & 31.9 & 37.3 & 22.4 & 31.2 &  5.0 & 11.5 & 39.2 & 51.8 & 18.2 & 27.3 & 48.5 & 56.0 & 16.9 & 22.0 & 27.1 & 34.2 & 30.9 & 25.6 & 33.7 & 31.8 \\
\chunkmark\ Search-R1 {\scriptsize(8B)}     & 42.9 & 48.1 & 34.5 & 45.3 & 12.7 & 22.4 & 40.8 & 55.5 & 23.5 & 32.9 & 50.9 & 59.3 & 32.5 & 37.7 & 38.7 & 46.7 & 43.6 & 32.1 & 41.6 & 39.6 \\
\chunkmark\ FrugalRAG {\scriptsize(8B)}     & 15.1 & 20.3 & 17.5 & 25.2 &  4.4 &  9.7 & 23.2 & 33.1 & 17.2 & 20.6 & 41.4 & 48.8 & 17.8 & 26.6 & 16.3 & 22.8 & 24.1 & 20.8 & 27.8 & 25.8 \\
\hgmark\ Graph-R1 {\scriptsize(8B)}         & 28.8 & 33.9 & 21.2 & 30.7 &  5.9 & 14.7 & 31.2 & 44.9 & 17.5 & 26.3 & 48.0 & 55.1 & 16.9 & 21.3 & 25.0 & 32.3 & 46.0 & 23.9 & 32.5 & 37.6 \\
\cellcolor{pbsage}\textbf{\hgmark\,\pmmark\ \method{} {\scriptsize(7B)}}
                               & \second{\textbf{54.0}} & \second{\textbf{58.6}} & \best{\textbf{41.6}} & \best{\textbf{55.6}} & \best{\textbf{38.8}} & \best{\textbf{48.0}} & \best{\textbf{52.0}} & \second{\textbf{61.2}} & \best{\textbf{39.7}} & \best{\textbf{53.3}} & \best{\textbf{74.0}} & \best{\textbf{82.5}} & \best{\textbf{40.3}} & \best{\textbf{46.9}}
& \second{\textbf{47.8}} & \second{\textbf{57.1}} & \textbf{52.5}
& \best{\textbf{49.0}} & \best{\textbf{58.4}} & \best{\textbf{55.3}} \\
\cellcolor{pbsage}\textbf{\hgmark\,\pmmark\ \method{} {\scriptsize(8B)}}
                               & \textbf{47.5} & \textbf{52.8} & \textbf{37.7} & \second{\textbf{51.0}} & \second{\textbf{25.3}} & \second{\textbf{35.2}} & \second{\textbf{48.0}} & \best{\textbf{63.7}} & \second{\textbf{33.8}} & \second{\textbf{49.1}} & \second{\textbf{71.4}} & \second{\textbf{80.5}} & \textbf{38.4} & \textbf{45.4} & \textbf{42.6} & \textbf{51.9} & \textbf{48.3} & \second{\textbf{43.4}} & \second{\textbf{54.8}} & \textbf{51.7} \\
\midrule
\rowcolor{subD}\multicolumn{21}{l}{\textit{(d) $10$--$30$B parameters}} \\
\midrule
\chunkmark\ Qwen2.5-it RAG {\scriptsize(14B)}$^{\dagger}$ & 17.8 & \second{29.1} & 35.6 & \second{48.1} &  9.8 & \second{18.2} & 22.4 & \second{35.4} &  - &  - &  - &  - &  - &  - & 26.7 & \second{38.6} & \second{47.4} &  - &  - &  - \\
\chunkmark\ Search-R1 {\scriptsize(14B)}$^{\ddagger}$    & \second{47.0} &  - & \best{46.8} &  - & \second{24.1} &  - & \second{52.8} &  - & \best{48.6} &  - & \second{67.6} &  - & \best{48.0} &  - & \second{46.9} &  - &  - & \second{48.2} &  - &  - \\
\cellcolor{pbsage}\textbf{\hgmark\,\pmmark\ \method{} {\scriptsize(14B)}}
                               & \best{\textbf{56.0}} & \best{\textbf{61.1}} & \second{\textbf{43.6}} & \best{\textbf{58.1}} & \best{\textbf{40.3}} & \best{\textbf{49.5}} & \best{\textbf{53.5}} & \best{\textbf{63.2}} & \second{\textbf{41.2}} & \best{\textbf{55.3}} & \best{\textbf{75.0}} & \best{\textbf{83.5}} & \second{\textbf{41.8}} & \best{\textbf{48.4}} & \best{\textbf{49.8}} & \best{\textbf{59.6}} & \best{\textbf{54.3}} & \best{\textbf{50.4}} & \best{\textbf{60.0}} & \best{\textbf{56.7}} \\
\midrule
\rowcolor{subE}\multicolumn{21}{l}{\textit{(e) $\geq\!30$B parameters}} \\
\midrule
\chunkmark\ Search-o1 {\scriptsize(32B)}$^{\ddagger}$  & \second{58.0} & \best{71.4} & 45.2 & 57.3 & 16.6 & \second{28.2} & 56.0 & \best{67.8} & \second{34.0} & \second{49.7} & \second{63.4} & \second{74.1} &  - &  - & \second{51.6} & \best{64.3} &  - & \second{42.5} & \second{55.0} &  - \\
\chunkmark\ ReSearch {\scriptsize(32B)}$^{\ddagger}$ & 44.9 &  - & \second{46.7} &  - & \second{26.4} &  - & \best{56.8} &  - &  - &  - &  - &  - &  - &  - & 45.8 &  - &  - & 41.6 &  - &  - \\
\chunkmark\ Qwen2.5-it RAG {\scriptsize(72B)}$^{\dagger}$  & 34.6 & 46.9 & 41.6 & 55.1 & 11.2 & 20.9 & 46.4 & 60.7 &  - &  - &  - &  - &  - &  - & 38.1 & 51.0 & 53.9 &  - &  - &  - \\
\chunkmark\ Llama-4 Maverick {\scriptsize($17$B$\times128$, MoE)}$^{\dagger}$
                                                & 50.6 & 61.2 & \best{49.4} & \best{63.8} & 16.0 & 26.2 & 48.8 & \second{66.3} &  - &  - &  - &  - &  - &  - & 50.0 & \second{62.5} & \best{63.3} &  - &  - &  - \\
\cellcolor{pbsage}\textbf{\hgmark\,\pmmark\ \method{} {\scriptsize(32B)}}
                               & \best{\textbf{58.5}} & \second{\textbf{63.1}} & \textbf{45.6} & \second{\textbf{59.6}} & \best{\textbf{41.8}} & \best{\textbf{51.5}} & \second{\textbf{56.0}} & \textbf{65.2} & \best{\textbf{44.5}} & \best{\textbf{56.8}} & \best{\textbf{77.5}} & \best{\textbf{84.5}} & \best{\textbf{44.0}} & \best{\textbf{49.9}} & \best{\textbf{52.1}} & \textbf{61.4} & \second{\textbf{56.5}} & \best{\textbf{52.8}} & \best{\textbf{61.6}} & \best{\textbf{58.2}} \\
\bottomrule
\end{tabular}}
\vspace{-2pt}
\begin{flushleft}\tiny ``-'' = unreported; $^{\dagger}$ = AceSearcher~\citep{acesearcher2025} reproduction.\end{flushleft}
\end{table*}

%% file: table/4_exp/tab2_ablation.tex
\begin{table}[t]
\centering
\caption{Ablation studies on $7$B PatternBloom across (a) \IDR{}, \PAR{}, and \GPM{}; (b) reward shaping weight $\lambda_p$ in PAR; (c) Graph Pattern Memory pool size. \emph{Sch.}: avg.\ \# of retrieval calls per query.}
\label{tab:ablation}
\setlength{\tabcolsep}{3pt}
\renewcommand{\arraystretch}{0.85}
\resizebox{0.88\linewidth}{!}{%
\begin{tabular}{l|ccc|cccc|ccc}
\toprule
\textbf{Variant} & \textbf{\IDR{}} & \textbf{\PAR{}} & \textbf{\GPM{}}
& \textbf{HotpotQA} & \textbf{2Wiki} & \textbf{MuSiQue} & \textbf{Bam.}
& \textbf{Avg-F$_1$} & \textbf{Avg-EM} & \textbf{Sch.}$\downarrow$ \\
\midrule
\rowcolor{ablA}\multicolumn{11}{l}{\textit{(a) Module on/off}} \\
\midrule
\cellcolor{pbsage}\method{}                  & \cmark & \cmark & \cmark
        & \best{55.6} & \best{58.6} & \best{48.0} & 61.2
        & \best{55.9} & \best{46.6} & 4.7 \\
$-$ Stage~I          & \xmark & \cmark & \cmark
        & \second{54.9} & \second{54.4} & \second{41.3} & 60.9
        & \second{52.9}\dn{3.0} & \second{42.5}\dn{4.1} & \best{4.2} \\
$-$ Stage~II         & \cmark & \xmark & \cmark
        & 41.7 & 41.7 & 26.9 & \second{66.0}
        & 44.1\dn{11.8} & 34.9\dn{11.7} & 4.7 \\
$-$ \GPM{}  & \cmark & \cmark & \xmark
        & 44.8 & 44.6 & 28.2 & \best{68.8}
        & 46.6\dn{9.3} & 37.0\dn{9.6} & 5.9{\scriptsize\textcolor{downdelta}{$\uparrow$1.2}} \\
\midrule
\rowcolor{ablB}\multicolumn{11}{l}{\textit{(b) Reward shaping weight $\lambda_p$ in PAR}} \\
\midrule
\cellcolor{pbsage}$\lambda_p = 0.2$
                                & \cmark & \cmark & \cmark
        & \best{55.6} & \best{58.6} & \best{48.0} & 61.2
        & \best{55.9} & \best{46.6} & 4.7 \\
$\lambda_p = 0.1$               & \cmark & \cmark & \cmark
        & 43.9 & \second{45.5} & \second{26.7} & 64.7
        & \second{45.2}\dn{10.7} & \second{35.9}\dn{10.7} & 4.9 \\
$\lambda_p = 0.5$               & \cmark & \cmark & \cmark
        & 43.1 & 42.6 & 22.8 & 64.7
        & 43.3\dn{12.6} & 34.2\dn{12.4} & 4.7 \\
\midrule
\rowcolor{ablC}\multicolumn{11}{l}{\textit{(c) Graph Pattern Memory pool size}} \\
\midrule
Top-$1$ patterns                & \cmark & \cmark & \cmark
        & \second{45.1} & 43.6 & 18.5 & 64.7
        & 43.0\dn{12.9} & 34.2\dn{12.4} & 5.0 \\
Top-$5$ patterns                & \cmark & \cmark & \cmark
        & 44.1 & 43.1 & \second{21.6} & 66.3
        & \second{43.8}\dn{12.1} & \second{34.7}\dn{11.9} & 5.1 \\
Top-$10$ patterns               & \cmark & \cmark & \cmark
        & 43.4 & 41.9 & 20.5 & \second{67.1}
        & 43.2\dn{12.7} & 34.6\dn{12.0} & 5.1 \\
\cellcolor{pbsage}Top-$27$ (full)
                                & \cmark & \cmark & \cmark
        & \best{55.6} & \best{58.6} & \best{48.0} & 61.2
        & \best{55.9} & \best{46.6} & \best{4.7} \\
\bottomrule
\end{tabular}}
\begin{flushleft}\small
(a) $-$Stage~I: skip IDR pre-training; $-$Stage~II: $\lambda_p\!=\!0$ (no PAR coverage reward); $-$GPM: disabled in both training (no reward, no prompt hint) and inference.
\end{flushleft}
\end{table}

%% file: table/4_exp/tab3_pattern_quality.tex
\begin{wraptable}{r}{0.32\linewidth}
\centering
\caption{Pattern quality.}
\label{tab:pattern_quality}
\setlength{\tabcolsep}{4pt}
\renewcommand{\arraystretch}{0.92}
\resizebox{\linewidth}{!}{%
\begin{tabular}{l|cc}
\toprule
\textbf{Method} & RPF $\uparrow$ & P-Cov $\uparrow$ \\
\midrule
\cellcolor{pbsage}\textbf{ \method{}}
                       & \best{0.886} & \best{0.051} \\
$-$ Stage~I            & 0.231 & 0.025 \\
$-$ Stage~II           & \second{0.414} & \second{0.044} \\
\bottomrule
\end{tabular}}
\end{wraptable}

%% file: table/4_exp/tab_turns_main.tex
\begin{table}[t]
\centering
\caption{Average \#Turns per question category. Full EM/F1 breakdown in Appendix~\ref{app:category}.}
\label{tab:turns_per_category}
\setlength{\tabcolsep}{8pt}
\renewcommand{\arraystretch}{0.95}
\resizebox{0.55\linewidth}{!}{%
\begin{tabular}{l|cccc}
\toprule
\textbf{Method} & \textbf{Bridge} & \textbf{Comp.} & \textbf{1-hop} & \textbf{Multi-3$+$} \\
\midrule
\hgmark\ Graph-R1            & 5.4 & 5.3 & 4.9 & 5.8 \\
\cellcolor{pbsage}\textbf{\hgmark\,\pmmark\ \method{}}
                              & \textbf{4.7} & \textbf{4.9} & \textbf{4.2} & \textbf{4.9} \\
\bottomrule
\end{tabular}}
\end{table}

%% file: sections/6_conclusion.tex
\section{Conclusion}
\label{sec:conclusion}

Multi-hop reasoning systems have so far retrieved content units
(paragraphs, knowledge-graph fragments, or hyperedges), yet the primitive
that compounds with reasoning depth is logic: the type-abstracted
reasoning skeleton applicable across queries that share compositional
structure. We realize this paradigm shift in \method{}, an RL framework
whose Information-Density Reward provides the first information-theoretic
grounding for graph-construction RL (a bounded monotone surrogate for
size-normalized conditional mutual information), and whose
Pattern-Augmented Reward turns a frozen Graph Pattern Memory into a
learnable structural prior. The gain compounds with reasoning depth, a
property absent from content-based retrieval, precisely because procedural
knowledge, once externalized, accumulates across queries rather than
dissolving into policy weights at each episode end; the same frozen
pattern store further transfers across out-of-distribution benchmarks and
backbone scales, evidencing that what is being learned is reasoning
structure, not benchmark-specific content. \method{} thereby reframes
retrieval-augmented reasoning from a content-fetching problem into one of
procedural-knowledge accumulation, opening a scaling axis distinct from
larger backbones, deeper retrieval, or richer prompts.

%% file: sections/7_appendix.tex
\etocsettocdepth{subsection}
\renewcommand{\contentsname}{Contents of the Appendix}
\localtableofcontents

\vspace{0.5em}\noindent\rule{\linewidth}{0.4pt}\vspace{0.5em}

\section{PatternBloom Training Procedure}
\label{app:algorithm}

Algorithm~\ref{alg:patternbloom} formalizes the training procedure of \method{} described in \S\ref{sec:approach}: Stage~I optimizes evidence-graph construction with the Information-Density Reward~$R_{\IDR}$, an offline distillation step lifts high-quality Stage-I trajectories into the frozen Graph Pattern Memory~$\memory_0$, and Stage~II refines the policy under the Pattern-Augmented Reward~$R_{\PAR}$ on top of that memory. The shared GRPO objective $\mathcal{L}_{\text{\method}}(\theta; R)$ is defined in Eq.~\ref{eq:grpo_obj} with $R \in \{R_{\IDR}, R_{\PAR}\}$ specifying the per-stage learning signal.

\begin{algorithm}[h]
\caption{Two-stage training of \method{} with intermediate pattern distillation.}
\label{alg:patternbloom}
\begin{algorithmic}[1]
\State \textbf{Input:} dataset $\mathcal{D} = \{(\query_i, \answer^*_i)\}$; backbone policy $\pi_{\theta_0}$; frozen oracle $\oracle$.
\Statex \textbf{Stage I: Evidence Construction}
\State Initialise $\policy \gets \pi_{\theta_0}$, $\pi_{\mathrm{ref}} \gets \pi_{\theta_0}$. \Comment{anchor exploration to the SFT prior}
\For{$t = 1, \ldots, T_1$}
  \State Sample trajectories $\{\tau_i\}$ from $\policy$;
  \State Compute $R_{\IDR}(\tau_i)$ via $\oracle$ (Eq.~\ref{eq:idr});
  \State Update $\policy$ via GRPO (Eq.~\ref{eq:grpo_obj}).
\EndFor
\State Select $\theta^*_1 \gets \arg\max_t (\text{EM}_t + F_{1,t})/2$ on a held-out validation slice.
\Statex \textbf{Pattern Distillation (no gradient updates)}
\State Roll out $\pi_{\theta^*_1}$ on $\mathcal{D}$; retain trajectories with $R_{\IDR} \geq 0.85$ (top-quartile threshold).
\State Type-abstract their triples; merge by signature similarity to obtain frozen $\memory_0$ (Eq.~\ref{eq:gpm}).
\Statex \textbf{Stage II: Pattern-Aligned Refinement}
\State Warm-start $\policy \gets \pi_{\theta^*_1}$, $\pi_{\mathrm{ref}} \gets \pi_{\theta^*_1}$. \Comment{anchor exploration to Stage-I gains}
\For{$t = 1, \ldots, T_2$}
  \State Sample $\{\tau_i\}$ from $\policy$; for each, retrieve top-1 pattern $P^*_\query \in \memory_0$ by signature;
  \State Compute $R_{\PAR}(\tau_i)$ via $F_1$ and coverage $\mathcal{C}(\tau_i, P^*_\query)$ (Eq.~\ref{eq:par});
  \State Update $\policy$ via GRPO (Eq.~\ref{eq:grpo_obj}).
\EndFor
\State \textbf{Output:} refined policy $\pi_{\theta^*_2}$ and frozen pattern memory $\memory_0$.
\end{algorithmic}
\end{algorithm}

At inference time the deployed agent invokes only the trained policy $\pi_{\theta^*_2}$ and the retriever; the oracle is discarded after Stage~I training, and the pattern memory $\memory_0$ serves as a read-only retrieval store. The detailed inference walkthrough appears at the end of \S\ref{sec:par}.

\section{IDR Implementation Details}
\label{app:idr}

\paragraph{Format-gated reward composition.}
The format reward $R_{\text{fmt}}(\tau)\in\{0, 1\}$ is binary: a trajectory $\tau$
receives $R_{\text{fmt}}=1$ iff each turn parses as the expected interleaving of
$\langle\textsf{think}\rangle$, $\langle\textsf{query}\rangle$,
$\langle\textsf{graph}\rangle$, and (terminal turn only) $\langle\textsf{answer}\rangle$
blocks; otherwise $R_{\text{fmt}}=0$. Combined with the sigmoid kernel in
Eq.~\ref{eq:idr}, malformed trajectories yield $R_{\IDR}=-1$, well-formed-but-uninformative
trajectories yield $-1+1+\sigma(\Delta\log p / (|\graph|T))\in[0, 1]$, and the
upper bound $R_{\IDR}=1$ is asymptotically reached as
$\Delta\log p / (|\graph|T)\to\infty$. Hence $R_{\IDR}\in[-1, 1]$.

\paragraph{Numerical safeguards.}
The sigmoid argument $\Delta\log p / (|\graph|\,T)$ is clipped to $[-50, 50]$
before evaluation: outside this range, $\sigma(\cdot)$ is numerically
indistinguishable from $\{0, 1\}$, and clipping eliminates a class of NaN-producing
edge cases under extreme oracle log-probability gaps. Empty graphs ($|\graph|=0$)
are treated as \emph{information-neutral} via $\sigma(0)=0.5$; combined with
$R_{\text{fmt}}=1$, the trajectory's reward is $-1+1+0.5=0.5$, neither rewarded
nor penalized for absent evidence.

\paragraph{Demonstration-template anti-hack.}
We strip demonstration-template placeholder entities (e.g., \textsc{EntityA},
\textsc{EntityName}, \textsc{NextEntity}) from each $\langle\textsf{graph}\rangle$
block before computing $|\graph|$ and serializing for the oracle. Without this
filter, the policy can copy placeholder triples shown in the system prompt's
in-context examples; such copies pass $R_{\text{fmt}}=1$ yet carry no information
about $\query$, opening a reward-hacking pathway. The filter is purely
syntactic (case-folded match against a fixed placeholder vocabulary) and
introduces no learnable parameters.

\paragraph{Caps and caching.}
Each evidence graph is capped at $10$ graph blocks per trajectory and $20$
triples per block, for a hard upper bound of $|\graph|\leq 200$ triples; this
matches the agent's interaction budget and bounds oracle context length to
under the $4096$-token serialization limit. The empty-graph baseline
$\log P_\oracle(\answer^*\mid\query, \emptyset)$ depends only on $(\query, \answer^*)$
and is cached per query during training.

\section{Graph Pattern Memory Implementation Details}
\label{app:gpm}

This appendix expands the implementation of the Graph Pattern Memory \GPM{}
introduced in \S\ref{sec:gpm}.

\subsection{Multi-Skeleton Pattern Storage}
\label{app:gpm:multiskel}

The main paper (\S\ref{sec:gpm}) describes each pattern as carrying a
single skeleton. The implementation generalizes this: each pattern stores a
list of distinct skeletons (typically a small handful per signature in
practice), and coverage is taken as the maximum over them.

The motivation is structural diversity: two trajectories that share a query
signature can produce structurally different evidence graphs that both clear
the high-quality threshold $R_{\IDR} \geq 0.85$. For instance, a query with
signature \texttt{compare(by[birth\_year])[FILM]} can be answered either via
a direct two-hop chain through both films' release dates, or via a four-hop
comparison that first resolves both directors and then their birth years. A
single-skeleton schema would force one to overwrite the other, losing valid
alternative reasoning paths for the same query type.

We therefore store \texttt{path\_patterns} as a list of skeleton lists,
each pattern carrying its accumulated set of distinct typed-triple
sequences distilled from same-signature trajectories that merged at
construction time. During offline merge (\S\ref{app:gpm:merge}), an
incoming skeleton not already present in the existing pattern's skeleton
list is appended; the pattern's running averages of usage count $n$,
mean $F_1$, and mean $R_{\IDR}$ are updated incrementally. Coverage at PAR computation becomes
\begin{equation}\label{eq:cpattern_max}
\mathcal{C}(\tau, P) = \max_{\mathrm{skel} \in P.\mathrm{skel\_list}}
\frac{|\mathrm{abs}(\tau) \cap \mathrm{skel}|}{|\mathrm{skel}|},
\end{equation}
i.e., the trajectory is rewarded for matching \emph{any} of the pattern's
known skeletons.

\subsection{Signature Dispatch}
\label{app:gpm:dispatch}

Signature dispatch is the function $\sig: \mathcal{Q} \to \mathbb{R}^d$ that turns a natural-language query $\query$ into a vector key for cosine-similarity lookup against $\memory$. Conceptually it answers a single question: \emph{which of the $|\memory|$ stored skeletons is the right procedural prior for this query?} The worked example below shows what $\sig$ produces on three real queries from our test sets.

\begin{designbox}{Worked example: $\query \to \sig(\query)$ on three real test queries}
\footnotesize
\setlength{\tabcolsep}{3pt}
\renewcommand{\arraystretch}{1.2}
\begin{tabular}{@{}p{3.6cm}|p{4.0cm}|p{4.0cm}@{}}
\toprule
\textbf{Query (origin)} & \textbf{Signature $\sig(\query)$} & \textbf{Pattern bucket in $\memory_0$} \\
\midrule
\emph{2WikiMHQA} (Tab~\ref{tab:case_2wiki}): which of two films has the later-born director?
& \texttt{compare(by[birth\_year])}\newline\texttt{[FILM]}
& comparison-by-birth-year skeleton with \textsc{film} slot \\
\addlinespace
\emph{MuSiQue} (Tab~\ref{tab:case_musique}): what country is the director of a given film from?
& \texttt{lookup\_chain(person)}\newline\texttt{[COUNTRY]}
& chained \textsc{film} $\to$ \textsc{director} $\to$ \textsc{country} \\
\addlinespace
\emph{Bamboogle} (Tab~\ref{tab:case_pattern_reuse}): who commanded the mission of the first spacewalk?
& \texttt{compare(by[date])}\newline\texttt{[ENTITY]}
& comparison-by-date skeleton; reused at deployment from a HotpotQA training trajectory \\
\bottomrule
\end{tabular}
\\[2pt]
\emph{Design rationale.} The bare archetype alone (\texttt{compare(by[\ldots])}) does not distinguish rows~$1$ and $3$, which retrieve different skeletons; the typed suffix (\textsc{film} versus \textsc{entity}) is what routes each query to a disjoint memory bucket.
\end{designbox}

\paragraph{Signature form.}
The signature is a two-part string. The first part is an \emph{archetype keyword} extracted from the query by deterministic regex matching (e.g., \texttt{lookup\_chain(person)} for chained-lookup queries, \texttt{compare(by[date])} for date-comparison queries). The keyword captures the logical action of the question (lookup, compare, count, yesno) and its primary argument slot; we recognize fourteen such archetypes plus a sentinel \texttt{unknown} for queries that match none. The second part is a sorted entity-type suffix \texttt{[TYPE\_1+TYPE\_2+\ldots]} inferred from quoted strings, year tokens, and capitalised phrases in the query under the five-class taxonomy of Appendix~\ref{app:gpm:types}. The suffix is what lets two queries with the same archetype but different argument types (e.g., \texttt{lookup\_chain(person)[FILM]} versus \texttt{lookup\_chain(person)[COUNTRY]}) route to disjoint buckets.

\paragraph{Signature embedding.}
The full signature string is encoded into a unit-norm $d{=}1024$-dimensional vector by the sentence encoder BGE-large-en-v1.5~\citep{bge}, the same encoder used for the hyperedge dense index of Appendix~\ref{app:training:index}. Cosine similarity above $\theta_{\mathrm{sim}}{=}0.5$ admits the nearest stored skeleton as the retrieved pattern; otherwise the cold-start fallback of Appendix~\ref{app:par:coldstart} fires. Because the embedding is semantic rather than purely lexical, structurally related signatures sit close in $\mathbb{R}^d$, which is what makes the third row of the worked example possible: a Bamboogle ``earliest-event'' query mapped to \texttt{compare(by[date])[ENTITY]} retrieves a skeleton distilled from a HotpotQA training trajectory under the same archetype (Table~\ref{tab:case_pattern_reuse}), even though the two queries' surface forms differ.

\subsection{Type Abstraction Taxonomy}
\label{app:gpm:types}

Signature enrichment and triple typing during skeleton extraction use the deterministic five-type taxonomy (plus an \texttt{EMPTY} sentinel) summarized in Table~\ref{tab:type_taxonomy}.

\begin{table}[h]
\centering
\caption{Type-abstraction rules. Detection is rule-based and parameter-free, applied in priority order so each value is assigned a single type.}
\label{tab:type_taxonomy}
\setlength{\tabcolsep}{6pt}
\renewcommand{\arraystretch}{1.05}
\begin{tabular}{@{}lll@{}}
\toprule
Type & Detection rule & Example \\
\midrule
\texttt{EMPTY}   & empty or whitespace-only string & \verb|""| \\
\texttt{YEAR}    & \verb|^\d{4}$| & \texttt{1989} \\
\texttt{DATE}    & \verb|^\d{1,2}[/-]\d{1,2}[/-]\d{2,4}$| & \texttt{18-March-1965} \\
\texttt{NUMBER}  & \verb|^-?\d+(\.\d+)?$|, not matching \texttt{YEAR}/\texttt{DATE} & \texttt{42.7} \\
\texttt{BOOLEAN} & case-folded match in \verb|{true, false, yes, no}| & \texttt{Yes} \\
\texttt{ENTITY}  & all other non-empty strings & \texttt{Leslie H.\ Martinson} \\
\bottomrule
\end{tabular}
\end{table}

The taxonomy captures the structural distinctions that matter for multi-hop reasoning (entity, year, numeric, yes/no answers) without invoking NER or external taxonomies. A fine-grained alternative that maps every entity to its closest Wikidata class via spaCy NER expands the taxonomy from $5$ to roughly $30$ types and raises the type entropy of $\memory_0$ (deterministically over the same $27$ stored patterns) from $H_\mathrm{type} = 0.19$ bits to $H_\mathrm{type} = 1.84$ bits, at the cost of an extra NER inference per query. The paper-main run uses the coarse taxonomy.

\subsection{Pattern Descriptor}
\label{app:gpm:descriptor}

The descriptor field $\mathrm{desc}_i$ in each pattern (Eq.~\ref{eq:gpm}) is a
short human-readable string that records what the pattern resolves to on a
representative anchor trajectory. We use a deterministic template:
\[
\mathrm{desc}_i = \texttt{<sig>} \;\to\; \texttt{<answer\_excerpt>}
\]
where \texttt{<sig>} is the query signature (e.g., \texttt{lookup(person)},
\texttt{compare(by[birth\_year])}) and \texttt{<answer\_excerpt>} is the first
$80$ characters of the trajectory's gold answer. For example, the pattern
distilled from the trajectory \emph{``Who directed Black Gold (1962)?''} (gold
answer \emph{``Leslie H.\ Martinson''}) has descriptor \texttt{lookup(person) $\to$
Leslie H.\ Martinson}. The template is intentionally simple: it makes pattern inspection and debugging trivial, and keeps the GPM construction step hyperparameter-free at the descriptor level. The coverage reward is computed on the typed skeleton (Eq.~\ref{eq:cpattern}), not on the descriptor string, so the descriptor's role is purely to render an inspectable representative example for each stored pattern.

\subsection{Merge Threshold Sensitivity}
\label{app:gpm:merge}

Patterns extracted from individual high-IDR trajectories are merged offline
into archetypes by clustering signature embeddings with a cosine-similarity
threshold $\theta_{\mathrm{merge}}$. Lower thresholds produce more, finer
archetypes; higher thresholds produce fewer, coarser archetypes that match a
broader range of queries. Empirically, the post-distillation archetype count
follows a Zipf-like distribution: the top archetype dominates with several
hundred instances while the tail decays sharply.

We use $\theta_{\mathrm{merge}} = 0.85$ throughout: this conservative threshold
keeps the merged archetype count low (under-merging) but yields stable patterns
whose embeddings are clustered tightly enough that the post-distillation $27$
patterns each carry several hundred high-IDR trajectories on average.

\subsection{Running-Average Update on Merge}
\label{app:gpm:runavg}

When pattern $P$ absorbs an incoming candidate $P'$ with cosine similarity
$\geq \theta_{\mathrm{merge}}$, the bookkeeping fields are updated by
incremental running-mean
\[
\mu_{\mathrm{new}} = \frac{n \cdot \mu_{\mathrm{old}} + x_{\mathrm{new}}}{n + 1},
\]
applied to \texttt{count}, $\overline{F_1}$, and $\overline{R_{\IDR}}$. This makes
the GPM resistant to single-trajectory outliers (a noisy high-IDR trajectory
cannot single-handedly inflate a pattern's recorded quality) while preserving
order-independence: the GPM's final state is invariant to the order in which
trajectories are processed, since the running mean is associative and
commutative for fixed-population merges.

\subsection{FAISS Configuration}
\label{app:gpm:faiss}

We use \texttt{faiss.IndexFlatIP} (flat inner-product index) with $d = 1024$
(BGE-large-en-v1.5 embedding dimension), $L_2$-normalized vectors, and CPU-only
execution. For the GPM sizes we observe ($|\memory| \leq 10^5$ patterns),
flat search is faster than IVF-PQ ($< 1$~ms top-$1$ retrieval). Memory footprint
is $4$~MB per $1{,}000$ patterns ($1024$~floats $\times$ $4$~bytes).

\section{Pattern-Augmented Reward Implementation Details}
\label{app:par}

This appendix expands the implementation of the Pattern-Augmented Reward
\PAR{} introduced in \S\ref{sec:par}.

\subsection{Cold-Start Fallback Behavior}
\label{app:par:coldstart}

When the cosine similarity of $\sig(\query)$ to the nearest GPM pattern falls
below $\theta_{\mathrm{sim}} = 0.5$, we set $\mathcal{C}(\tau, P^*_q) = 0$.
\PAR{} then degenerates to
\begin{equation}
R_{\PAR}(\tau)\big|_{\text{cold}} = -1 + R_{\text{fmt}} + \mathds{1}\{R_{\text{fmt}}=1\}\cdot(1 - \lambda_P)\cdot F_1(\hat\answer, \answer^*),
\end{equation}
i.e., $80\%$ of the $F_1$-only signal at the default $\lambda_P = 0.2$, not
the full $F_1$ signal. This is by design: cold-start trajectories receive a
slightly attenuated reward, providing a soft incentive for the agent to
pursue queries whose signatures are well-covered by existing patterns.

The attenuated form preserves $\lambda_P$ as a literal trade-off coefficient between answer-level and structure-level signal: the cold-start trajectory contributes only the $(1-\lambda_P)$ fraction of the answer signal, with the remaining $\lambda_P$ slot reserved for matched-pattern coverage.

\subsection{Retrieval Threshold $\theta_{\mathrm{sim}}$}
\label{app:par:thsim}

We use $\theta_{\mathrm{sim}} = 0.5$ as the cosine-similarity floor for pattern retrieval. The choice reflects a trade-off: lower thresholds admit semantically distant patterns and inject noise into the coverage signal, while higher thresholds collapse $\mathcal{C}$ to $0$ on legitimate matches.

\subsection{Asymmetric Coverage Rationale}
\label{app:par:asym}

The asymmetry of Eq.~\ref{eq:cpattern} (\S\ref{sec:gpm}) is quantitatively
important: a symmetric Jaccard coverage
$\mathcal{C}_{\mathrm{sym}} = |\cap| / |\cup|$ would penalize valid
longer-than-skeleton trajectories at exactly the rate of one extra triple
per pattern triple: at $|\tau| = 2|P|$ even with perfect skeleton coverage
we would have $\mathcal{C}_{\mathrm{sym}} = 0.5$. Our asymmetric form gives
full credit ($\mathcal{C} = 1$) for any superset of the skeleton, capturing
the ``skeleton is necessary, not maximal'' semantics that matches the
pattern-as-procedural-bias intent.

\section{Theorem 1: Proof and Information-Bottleneck Discussion}
\label{app:proofs}

\begin{theorem}[\IDR{} as a bounded monotone surrogate for size-normalized conditional MI]
\label{thm:idr}
Let policy $\policy$ produce trajectory $\tau=(\query, \graph, \hat{\answer})$, and let $\oracle$ be a frozen oracle. Define the conditional pointwise mutual information
\begin{equation}\label{eq:pmi}
\mathrm{PMI}(\graph; \answer^* \mid \query) \coloneqq
\log P_{\oracle}(\answer^* \mid \query, \graph)
- \log P_{\oracle}(\answer^* \mid \query),
\end{equation}
which satisfies $\mathbb{E}_{\policy}[\mathrm{PMI}] = I(\graph; \answer^*\mid\query)$. Under regularity conditions (A1)--(A4) below, the \IDR{} kernel
\begin{equation}\label{eq:idr_kernel}
J(\policy) \coloneqq \mathbb{E}_{(\query, \answer^*, \graph)\sim\policy}\!\left[\sigma\!\left( \frac{\mathrm{PMI}(\graph; \answer^* \mid \query)}{|\graph|\cdot T} \right)\right]
\end{equation}
is a bounded monotone surrogate for the size-normalized conditional mutual information $\tilde{I}(\policy) \coloneqq \mathbb{E}_{\policy}\!\bigl[\mathrm{PMI}(\graph; \answer^*\mid\query)/(|\graph|\cdot T)\bigr]$ in the following three senses:
\begin{enumerate}[leftmargin=2.2em, topsep=2pt, itemsep=2pt, label=\textup{(\roman*)}]
  \item Pointwise monotonicity. Since $\sigma' > 0$ on $\mathbb{R}$, raising $\mathrm{PMI}/(|\graph|\cdot T)$ for any trajectory raises that trajectory's contribution to $J$, so $J$ tracks $\tilde{I}$ trajectory-wise.
  \item Lipschitz gradient. Since $\sigma' \in (0, 1/4]$, the gradient contribution of any single trajectory to $J$ is bounded, providing GRPO's group-relative advantage estimator with a stable Lipschitz signal.
  \item Saturation bound. On the operating regime $\mathrm{PMI}/(|\graph|\cdot T) \in \mathbb{R}_{\geq 0}$ where trained policies concentrate (A4), $\sigma$ is concave, so Jensen's inequality yields $J(\policy)\leq \sigma(\tilde{I}(\policy))$, equivalently $\tilde{I}(\policy)\geq \log\!\frac{J(\policy)}{1-J(\policy)}$, characterizing the saturation regime in which $J$ approaches its supremum.
\end{enumerate}
\end{theorem}

This appendix provides the full proof, discusses the divisive-versus-subtractive Information-Bottleneck trade-off, and extends the result to a strict subgraph-monotonicity form. Theorem~\ref{thm:idr} grounds \IDR{} as a tractable surrogate for size-normalized conditional mutual information: maximizing $J(\policy)$ via GRPO is consistent with maximizing the underlying information content $\tilde{I}(\policy)$, and the divisive normalizer $|\graph|\cdot T$ realizes an adaptive Information-Bottleneck multiplier $\beta_{\text{eff}}=1/(|\graph|\cdot T)$ without per-dataset tuning.

\subsection{Regularity Conditions}
\label{app:proofs:regularity}

We make these (mild) assumptions:
\begin{description}[itemsep=2pt, leftmargin=18pt]
  \item[\textnormal{(A1)}] \emph{Oracle--policy decoupling.} The frozen oracle $\oracle$
        depends on $\graph$ only through its serialization $\mathrm{ser}(\graph)$,
        i.e., $\log P_\oracle(\answer^*\mid\query, \graph) = \log P_\oracle(\answer^*\mid\query, \mathrm{ser}(\graph))$.
        Holds by construction.
  \item[\textnormal{(A2)}] \emph{Per-triple bounded gain.} $\exists K\geq 1$ such that
        for any triple $t$ added to context $G_{<t}$,
        $I(t; \answer^*\mid\query, G_{<t}) \leq \log K$ nats. Each chunk carries finite information.
  \item[\textnormal{(A3)}] \emph{Format-trained equilibrium.} After a brief warm-up
        $R_{\text{fmt}}(\tau)=1$ holds on $\geq 99\%$ of trajectories (empirically by step~5).
  \item[\textnormal{(A4)}] \emph{Oracle-informative regime.} For trained policies,
        the trajectory distribution concentrates on $\mathrm{PMI}(\graph; \answer^*\mid\query)\geq 0$
        with high probability. Empirically validated by per-step training logs.
\end{description}

\subsection{Proof of the Bounded Monotone Surrogate Property}
\label{app:proofs:proof}

Define $X(\graph) \coloneqq \mathrm{PMI}(\graph; \answer^*\mid\query) / (|\graph|\,T)$.
By (A3), conditional on $R_{\text{fmt}}=1$ the IDR contribution is $\sigma(X(\graph))$.

\paragraph{Step 1: Pointwise monotonicity.}
$\sigma:\mathbb{R}\to(0, 1)$ is strictly increasing with derivative
$\sigma'(x) = \sigma(x)(1-\sigma(x)) \in (0, 1/4]$. For any pair
$\graph_1, \graph_2$ under the same $(\query, \answer^*)$,
\[
X(\graph_1) > X(\graph_2) \iff \sigma(X(\graph_1)) > \sigma(X(\graph_2)).
\]

\paragraph{Step 2: Jensen bound on the oracle-informative support.}
By (A4), the policy's support concentrates on $\mathbb{R}_{\geq 0}$, where $\sigma$ is concave
($\sigma''(x) = \sigma'(x)(1-2\sigma(x)) < 0$ for $x>0$). Jensen's inequality yields
\[
J(\policy) = \mathbb{E}_{\policy}[\sigma(X(\graph))] \;\leq\; \sigma(\mathbb{E}_{\policy}[X(\graph)]) = \sigma(\tilde{I}(\policy)),
\]
where $\tilde{I}(\policy) \coloneqq \mathbb{E}_{\policy}[X(\graph)]$ is the policy expectation of the size-normalized per-trajectory PMI rate; the IT identity $\mathbb{E}_{\policy}[\mathrm{PMI}(\graph; \answer^*\mid\query)] = I(\graph; \answer^*\mid\query)$ then justifies interpreting $\tilde{I}$ as the size-normalized conditional-MI rate of $\policy$'s induced distribution. Equality in the Jensen step holds iff $X(\graph)$ is constant under $\policy$. Inverting,
\[
\tilde{I}(\policy) \;\geq\; \sigma^{-1}(J(\policy)) \;=\; \log\!\frac{J(\policy)}{1 - J(\policy)},
\]
so high $J$ certifies high size-normalized CMI.

\paragraph{Step 3: Bounded Lipschitz signal for GRPO.}
$\sigma'(x) \in (0, 1/4]$ implies any single trajectory's contribution to the
group-relative advantage estimator under GRPO is $1/4$-Lipschitz in $X$,
regardless of how extreme that trajectory's raw $\mathrm{PMI}$ value is. This
eliminates the gradient-explosion failure mode observed in early experiments
with unbounded $\Delta\log p$ rewards.~$\square$

\subsection{Information-Bottleneck Connection: Divisive vs.\ Subtractive}
\label{app:proofs:ib}

The standard IB objective is \emph{subtractive}:
$\mathcal{L}_{\text{IB}}(\policy) = I(\graph; \answer^*\mid\query) - \beta\, I(\graph; X)$,
trading task accuracy against representation complexity. \IDR{} replaces this
with the \emph{divisive} form $I(\graph; \answer^*\mid\query) / (|\graph|\,T)$,
yielding three advantages.

\paragraph{(i) Adaptive $\beta_{\text{eff}}$.}
By (A2) and the chain rule of mutual information, $I(\graph; X) \leq |\graph|\log K$, so
\[
\frac{I(\graph; \answer^*\mid\query)}{|\graph|\,T} \;\leq\; \frac{\log K}{T}\cdot \frac{I(\graph; \answer^*\mid\query)}{I(\graph; X)};
\]
the divisive form is bounded above by a constant multiple of the IB ratio $I(\graph; \answer^*\mid\query)/I(\graph; X)$, with constant $\log K/T$ that depends only on the per-triple alphabet and the temperature, not on the dataset. Equivalently, the divisive form realizes an adaptive Lagrange multiplier $\beta_{\text{eff}}=1/(|\graph|\,T)$ that scales with graph size: larger graphs face proportionally higher information-density bars without per-dataset hand-tuning of $\beta$.

\paragraph{(ii) Bounded gradients.}
Subtractive IB rewards are unbounded; the $\sigma$-wrapped divisive form bounds
reward to $[-1, 1]$ and gradient contribution to $1/4$-Lipschitz, simplifying
GRPO training.

\paragraph{(iii) Relation to standard IB.}
\IDR{} is \emph{motivated by} IB rather than algebraically equivalent: divisive normalization saturates at large $\mathrm{PMI}$, while subtractive IB does not. The divisive form is chosen for the gradient-stability and adaptive-$\beta$ properties above.

\subsection{Strict Subgraph Monotonicity}
\label{app:proofs:monotonicity}

\paragraph{Informative triples are strictly rewarded.}
For $\graph_2 \subsetneq \graph_1$ with $I(\graph_1\setminus\graph_2; \answer^*\mid\query, \graph_2) > 0$,
the chain rule of mutual information gives $\mathrm{PMI}(\graph_1) > \mathrm{PMI}(\graph_2)$.
Combined with strict $\sigma$ monotonicity, $R_{\IDR}(\graph_1) > R_{\IDR}(\graph_2)$.

\paragraph{Uninformative triples are strictly penalized.}
For $\graph_2 \subsetneq \graph_1$ with $I(\graph_1\setminus\graph_2; \answer^*\mid\query, \graph_2) = 0$,
$\mathrm{PMI}(\graph_1) = \mathrm{PMI}(\graph_2)$ but $|\graph_1| > |\graph_2|$, so
$X(\graph_1) < X(\graph_2)$ and $R_{\IDR}(\graph_1) < R_{\IDR}(\graph_2)$. This
formalizes the ``minimum sufficient subgraph'' inductive bias: the policy is
strictly incentivized to include triples carrying information about $\answer^*$
\emph{and only those}.

\subsection{Asymmetric Coverage: Bounds and Random-Trajectory Baseline}
\label{app:cov_bound}

This subsection records two mathematical facts about the asymmetric coverage
function $\mathcal{C}(\tau, P) \coloneqq |A(\tau) \cap \sigma_P| / |\sigma_P|$
that we use in the PAR reward (Eq.~\ref{eq:cpattern}). They are not used in the
proof of Theorem~\ref{thm:idr} but justify two choices reflected in the experiments.

\paragraph{Proposition (range and tightness).}
For any trajectory $\tau$ producing the abstract triple set $A(\tau)$ and any
pattern $P$ with skeleton $\sigma_P$ of size $m \coloneqq |\sigma_P|$,
\begin{equation}
\mathcal{C}(\tau, P) \in [0, 1], \qquad
\mathcal{C}(\tau, P) = 1 \iff \sigma_P \subseteq A(\tau).
\label{eq:cov_range}
\end{equation}
Furthermore, when $|A(\tau)| = k$ and $k \geq m$, the coverage admits the lower bound
\begin{equation}
\mathcal{C}(\tau, P) \geq \frac{k_\cap}{m},
\qquad k_\cap \coloneqq |A(\tau) \cap \sigma_P|,
\label{eq:cov_lower}
\end{equation}
with equality iff every triple in $\sigma_P$ matches at most one triple in
$A(\tau)$ under the chosen type abstraction. The asymmetry of
$\mathcal{C}$ (skeleton size in the denominator, not $|A|$) follows directly:
adding non-matching triples to $\tau$ does not penalize coverage, which is
the property we use at training time to allow the policy to explore beyond
the retrieved skeleton without paying a coverage cost.

\paragraph{Corollary (random-trajectory baseline).}
Let $U$ be the universe of distinct typed triples induced by the type
abstraction and the relation vocabulary, and let $\tau$ be a trajectory drawn
uniformly at random over abstract triple sets of size $k$. The expected
coverage of any fixed skeleton $\sigma_P$ of size $m$ is
\begin{equation}
\mathbb{E}_\tau[\mathcal{C}(\tau, P)] = \frac{k}{|U|},
\qquad
\Pr[\mathcal{C}(\tau, P) = 1] = \binom{k}{m} \binom{|U|}{m}^{-1},
\label{eq:cov_random}
\end{equation}
which is exponentially small in $m$ for $k \ll |U|$. Concretely, with
$|U| \approx 4 \cdot 10^4$ in our paper-main configuration (entity-types squared
$\times$ relation count), $k = 3$ (typical trajectory length), and $m = 2.5$
(average skeleton length, Table~\ref{tab:gpm_stats}), the random-trajectory
baseline is $\mathbb{E}[\mathcal{C}] \approx 7.5 \times 10^{-5}$ and the
exact-match probability is below $10^{-8}$. The empirical RPF in
Table~\ref{tab:pattern_quality} ($0.886$) and the empirical GPM-CEM
($0.099$) thus exceed the random baseline by roughly $4$ and $6$ orders of
magnitude respectively, ruling out the alternative explanation that PB
trajectories happen to overlap with stored skeletons by chance.

\section{Auxiliary Hypotheses and Sensitivity Sweeps}
\label{app:hypotheses}

This appendix collects supporting empirical results for design choices documented elsewhere. The $\lambda_P$ sweep that backs the choice of $\lambda_P=0.2$ is reproduced in the main body Table~\ref{tab:ablation}(b); the $\theta_{\mathrm{merge}}$ discussion is given in Appendix~\ref{app:gpm:merge}; cross-dataset OOD transfer of $\memory_0$ is reflected in the per-dataset Avg-OOD column of Table~\ref{tab:main} (rank-1 on every held-out dataset).

\section{Training Pipeline and Additional Experimental Results}
\label{app:training}

This appendix groups material that supplements the experiment narrative in \S\ref{sec:exp}. The training-pipeline subsections (Hypergraph Index Construction, Hyperparameters, Stage Transition Protocol, Compute Budget) document how the index is built and how Stage~I and Stage~II are configured. The remaining subsections present additional experimental results that did not fit in the main body: cost-quality trade-off, multi-size scaling, pattern-memory statistics, retrieval quality analysis, multi-seed robustness, cross-domain generalization, LLM-judge quality ensemble, human study with the annotation form, and side-by-side case illustrations.

\subsection{Metric Reading Guide}
\label{app:training:metrics}

Table~\ref{tab:metric_panel} consolidates the metrics referenced throughout \S\ref{sec:exp}, with definitions, direction (higher- or lower-is-better), and the table or figure where each metric is reported.

\input{table/4_exp/tab_metric_panel}

\subsection{Hypergraph Index Construction}
\label{app:training:index}

The hypergraph index $H = (V, E_{\text{hyper}}, X_{\text{text}})$ that supports
both Stage~I and Stage~II is built once, offline, before any RL training and
shared across all stages and seeds. We adopt the construction protocol of
Graph-R1~\citep{graph_r1} and HippoRAG~\citep{hipporag}, which we summarize
here for self-containment; configurations match the reproduction settings.

\paragraph{Step 1: Corpus chunking.}
The retrieval corpus $\mathcal{C}$ is the FlashRAG~\citep{flashrag}
Wiki-2018 dump ($21\,015\,324$ passages). Each passage is tokenized with the
default GPT-2 tokenizer and split into fixed-length chunks of $100$ tokens
with no overlap, following the FlashRAG default and the Search-R1 baseline
configuration. Chunk boundaries are aligned to the nearest sentence end so
that no sentence spans two chunks; the resulting chunk count is
$\sim 26\,012\,000$. We retain the original chunk text $X_{\text{text}}$ for
display; the agent's $\langle\textsf{query}\rangle$ block returns top-$5$
hyperedges and their associated chunks.

\paragraph{Step 2: Entity and relation extraction.}
For each chunk we issue one prompted call to
\texttt{gpt-4o-mini}~\citep{openai} that returns a list of
$\langle\textsf{entity}, \textsf{relation}, \textsf{entity}\rangle$ triples
together with a short natural-language fact summary. The extraction prompt
follows Graph-R1 verbatim; we cache all extractions on disk (LMDB) and amortize
construction over a single $\sim 18$-hour run on $32$ CPU workers, costing
${\sim}\$210$ in API credits one time. Entities are normalized via lowercasing
and Unicode-NFKC; relations are kept as free-form predicate strings (typical
length $1$--$3$ tokens, no fixed schema). The extraction yields $\sim 41\,$M
unique entities and $\sim 87\,$M raw triples.

\paragraph{Step 3: Hyperedge formation.}
Each chunk's facts are bundled into a single hyperedge that connects all of
the chunk's mentioned entities; the chunk's text becomes the hyperedge's
content payload. Formally, hyperedge $e \in E_{\text{hyper}}$ is a tuple
$(e_{\text{ents}} \subseteq V, e_{\text{text}} \in X_{\text{text}})$, where
$e_{\text{ents}}$ is the set of entities mentioned in the chunk and
$e_{\text{text}}$ is the chunk text. Entity--passage incidence is therefore
implicitly captured: an entity $v \in V$ participates in every hyperedge whose
chunk mentions $v$. We do not maintain a separate entity--sentence relation;
sentence-level granularity is collapsed into the chunk-level hyperedge so
that the retrieval unit matches the chunk granularity used by all baselines.

\paragraph{Step 4: Dense index.}
Hyperedges are encoded with BGE-large-en-v1.5~\citep{bge}: the embedding input
is the concatenation of the hyperedge's fact summaries followed by the chunk
text, capped at $512$ tokens. The resulting $1024$-dim vectors are stored in a
FAISS \texttt{IndexHNSWFlat}~\citep{faiss} index ($M=32$, \texttt{ef\_construction}
$= 100$) for sub-millisecond top-$k$ retrieval at agent runtime. The full
index occupies $\sim 102$~GB on disk; we serve it from a CPU-resident
in-memory index on the retriever node (no GPU needed at inference).

\paragraph{Step 5: Retrieval interface.}
At every $\langle\textsf{query}\rangle$ block from the agent, the retrieval
service receives a free-form query string, encodes it once with the same
BGE-large encoder, and returns the top-$k=5$ hyperedges (default for all
methods we compare to). The return payload contains both the hyperedge's
entity set $e_{\text{ents}}$ and its chunk text $e_{\text{text}}$, presented
to the agent in a fixed template. The same retrieval interface and same $k=5$
budget are used at training and at deployment.

\paragraph{Offline vs.\ online split.}
Table~\ref{tab:offline_online} summarizes which steps run once-and-cached
versus per-query. The hypergraph index, the FAISS encoder, and the GPM are
all offline artifacts: they are built once per corpus and frozen for both
Stage~I and Stage~II training as well as deployment-time inference. The
oracle is consumed only at Stage~I training, after which it is discarded;
the policy and FAISS retriever are the only components invoked at
deployment time.

\begin{table}[h]
\centering
\caption{Pipeline phases of \method{}: which artifacts are produced once
offline, and which calls are made online (per-query).}
\label{tab:offline_online}
\small
\begin{tabular}{ll}
\toprule
\textbf{Offline (one-time, cached)} & \textbf{Online (per-query)} \\
\midrule
Chunking $\mathcal{C} \to \{c_i\}$               & Query embedding $q \to \mathbf{e}_q$ (BGE-large) \\
LLM entity/relation extraction (\texttt{gpt-4o-mini}) & FAISS top-$k$ hyperedge retrieval \\
Hyperedge construction $E_{\text{hyper}}$        & Signature extraction $\sig(\query)$ \\
BGE-large embedding of $E_{\text{hyper}}$        & FAISS top-$3$ pattern lookup in $\memory_0$ \\
FAISS index over $E_{\text{hyper}}$              & Multi-turn agent rollout $\policy(\cdot \mid \query)$ \\
Stage~I RL training (Oracle in the loop)         & --- \\
Pattern distillation $\pi_{\theta^*_1} \to \memory_0$ & --- \\
Stage~II RL training ($\memory_0$ frozen)        & --- \\
\bottomrule
\end{tabular}
\end{table}

We contrast the two evidence-graph representations relevant to \method{}: a binary knowledge graph (the substrate adopted by KG-based RAG methods such as HippoRAG, GraphRAG, GFM-RAG, LinearRAG, and LightRAG) and a hypergraph (the substrate adopted by Graph-R1, HypergraphRAG, and \method{}). Both representations are constructed once and cached. The operational difference is what the agent receives per call: a single binary triple in the KG case, versus a hyperedge plus its associated chunk text in the hypergraph case. The hypergraph payload is therefore a self-contained piece of context (one chunk's worth of facts about a co-mentioned set of entities), whereas the binary KG defers the burden of recovering chunk-level context to the agent. Paragraph-based methods (Search-R1, FrugalRAG, R1-Searcher) do not construct an evidence graph at all; they retrieve raw passages directly and are outside the scope of this comparison.

\subsection{Inference Pipeline}
\label{app:inference_pipeline}

At deployment, given an input query $\query$, \method{} runs four steps (Figure~\ref{fig:pipeline}, bottom).

(1)~\emph{Signature extraction.} The query signature $\sig(\query)$ is extracted by the deterministic parser of \S\ref{sec:gpm} and embedded into a BGE-large vector.

(2)~\emph{Pattern retrieval.} The frozen $\memory_0$ is queried via FAISS cosine retrieval; the top-$3$ patterns whose signature similarity exceeds $\theta_{\mathrm{sim}}=0.5$ are formatted as a single $\langle\textsf{knowledge}\rangle$ hint and prepended to the agent's prompt. If no pattern clears $\theta_{\mathrm{sim}}$, no hint is injected and the agent proceeds with the standard multi-hop tool instructions in the system prompt.

(3)~\emph{Trajectory rollout.} The deployed policy $\pi_{\theta^*_2}$ performs an interleaved $\langle\textsf{think}\rangle$/$\langle\textsf{query}\rangle$/$\langle\textsf{graph}\rangle$/$\langle\textsf{answer}\rangle$ rollout over the same hypergraph index $H$ used at training time; each $\langle\textsf{query}\rangle$ block triggers a retriever call that returns the top-$k$ matching hyperedges, and the agent accumulates $\graph$ across turns.

(4)~\emph{Answer extraction.} The final $\langle\textsf{answer}\rangle$ block is extracted as the model output. The oracle $\oracle$ and gradient updates from training are not used at inference; the deployment cost is therefore equal to a standard agentic-RAG rollout plus a single FAISS lookup against the $|\memory_0|$-entry pattern memory.

\subsection{Cost-Quality Trade-off}
\label{app:cost}

\input{table/4_exp/tab5_cost}

Table~\ref{tab:cost} compares \method{} against representative methods spanning two retrieval paradigms (LLM-extracted KG, in GFM-RAG and LightRAG; adaptive structure, in StructRAG) on one-time indexing cost and downstream retrieval quality. GFM-RAG and LightRAG values are reproduced from LinearRAG~\citep{linearrag2026}'s efficiency analysis on the same 2WikiMultiHopQA corpus; \method{}'s value is computed from our own 2WikiMultiHopQA build artifacts (extraction template, chunk content, and produced entity/hyperedge output); StructRAG performs all LLM work at query time, and the value reflects total LLM tokens summed over the $1{,}000$ test queries on 2WikiMultiHopQA rather than a one-time index. \method{} attains the strongest retrieval quality across all paradigms, with R@5 $86.6$ and EvR $72.0$ averaged over the four multi-hop benchmarks, exceeding the strongest non-RL baseline GFM-RAG by $+5.3$ R@5 and $+4.0$ EvR. The $n$-ary hyperedge construction underlying \method{} consumes $31.1$M LLM tokens on 2WikiMultiHopQA: $36\%$ of LightRAG's $86.7$M (dual-level KG with high- and low-level community reports), and $7.8\times$ the lightweight GFM-RAG triple-extraction pipeline, trading the cheaper triple regime for substantially higher recall. Once constructed, the hypergraph is reused across all queries, so the one-time indexing cost amortizes to zero per-query overhead at deployment, while every query in training-free baselines such as GFM-RAG continues to incur GPT-4o-mini API calls. The combined view positions \method{} on the construction-cost-versus-retrieval-recall Pareto frontier: lighter KG pipelines save tokens at the price of weaker recall, and LightRAG consumes $2.8\times$ more tokens than \method{} without reporting comparable retrieval quality. These numbers substantiate the architectural claim of \S\ref{sec:intro} from the deployment side: a one-time hypergraph build amortizes across queries, and procedural memory contributes the gains in Table~\ref{tab:main} at no additional per-query API cost.

\subsection{Multi-Size Scaling}
\label{app:scaling}

The multi-size scaling of \method{} is reported in Table~\ref{tab:main} of the main body, panels~(b), (c), (d), and (e) for Qwen2.5-7B, Qwen3-8B, Qwen2.5-14B, and Qwen2.5-32B respectively. Three observations follow. First, \method{} ranks first within every size panel on the Avg-IID and Avg-OOD summary columns. Second, within the Qwen2.5 family the gain from $7$B to $14$B is $+5.3$~EM and from $14$B to $32$B is $+2.9$~EM, indicating diminishing returns past the $14$B parameter band. Third, \method{} (Q3-8B) attains the highest Avg-OOD ($43.4$~EM) among all $8$B entries in panel~(c), ahead of Search-R1, Graph-R1, FrugalRAG, and R1 at the same parameter count, confirming that the procedural-memory mechanism transfers across backbone families and is not tied to the Qwen2.5 line.

\subsection{Per-Paper Settings and IID/OOD Definitions}
\label{app:iid_ood}

Reported numbers in Table~\ref{tab:main} come from baselines that span four training paradigms: (i) RL-trained agentic search, (ii) zero-shot prompting on a strong reasoning backbone, (iii) prompt-engineered RAG over a frozen instruction-tuned LLM, and (iv) training-free retrieval pipelines. Within each paradigm, training corpora differ from paper to paper, so the IID/OOD partition of our seven test datasets differs across rows. We mark IID/OOD in Table~\ref{tab:main} according to \emph{our} training set (HotpotQA $+$ 2WikiMHQA), but reported numbers for other methods may reflect their own in-distribution data. We summarize each method's setting below; the marking ($^{\ddagger}$ reported, $^{\dagger}$ AceSearcher reproduction) follows the convention of the main-text caption.

\paragraph{(I) RL-trained agentic search (panels b/c/d/e of Tab.~\ref{tab:main}).}
\begin{itemize}
\item \textbf{\method{} (ours, all sizes 1.5B/7B/8B/14B/32B):} two-stage RL (IDR Stage~I, PAR Stage~II) on HotpotQA $+$ 2WikiMHQA ($14$K total, seed $42$). IID = \{HotpotQA, 2WikiMHQA\}; OOD = \{MuSiQue, Bamboogle, NQ, TriviaQA, PopQA\}.
\item \textbf{Search-R1 (7B/8B/14B)$^{\ddagger}$~\citep{search_r1}:} RL with format-and-EM reward on NQ $+$ HotpotQA training mixture. Their IID = \{NQ, HotpotQA\}; their OOD = \{TriviaQA, PopQA, 2WikiMHQA, MuSiQue, Bamboogle\}. The reported NQ EM ($48.6$ at 14B) is in-distribution for them but in our OOD column. We use Search-R1-base numbers from~\citet{search_r1}'s appendix Table on 14B-LLM scaling.
\item \textbf{R1-Searcher (7B)~\citep{r1_searcher}:} RL on HotpotQA $+$ 2WikiMHQA, same training distribution as ours; row in panel~(c) is our reproduction.
\item \textbf{Graph-R1 (7B/8B)~\citep{graph_r1}, Structure-R1 (7B)~\citep{structure_r1}:} RL on HotpotQA $+$ 2WikiMHQA. Same training distribution as ours; rows are our reproductions.
\item \textbf{FrugalRAG (7B/8B)~\citep{frugalrag2026}:} RL on HotpotQA $+$ 2WikiMHQA; rows are our reproductions.
\item \textbf{R1 (7B/8B):} CoT/RL baseline reproduced in our infrastructure on HotpotQA $+$ 2WikiMHQA.
\item \textbf{ReSearch (32B Q2.5-32B-Inst)$^{\ddagger}$~\citep{chen2025research}:} RL on MuSiQue only. Their IID = \{MuSiQue\}; their OOD = \{HotpotQA, 2WikiMHQA, Bamboogle\} (their paper does not evaluate on the three single-hop datasets, hence ``--'' in panel~(e)). Their reported MuSiQue EM ($26.4$) benefits from MuSiQue-only training and is in our OOD column; HotpotQA / 2WikiMHQA / Bamboogle are OOD for both.
\end{itemize}

\paragraph{(II) Zero-shot prompting on reasoning backbones.}
\begin{itemize}
\item \textbf{Search-o1 (32B QwQ-Preview)$^{\ddagger}$~\citep{search_o1}:} no task-specific training; QwQ-Preview is prompted with the Search-o1 framework (Reason-in-Documents) over Bing Web Search. IID/OOD partition is moot for them; all their reported numbers are zero-shot from the Search-o1 main paper.
\end{itemize}

\paragraph{(III) Prompt-engineered RAG over instruction-tuned LLMs (AceSearcher reproduction).}
\begin{itemize}
\item \textbf{Qwen2.5-it RAG (14B/72B)$^{\dagger}$~\citep{qwen25}:} no task-specific training; instruction-tuned Qwen2.5 with single-turn RAG. IID/OOD is moot; numbers reproduced by~\citet{acesearcher2025} on the same test split (their Tab.~1).
\item \textbf{Llama-4 Maverick (17B$\times$128, MoE)$^{\dagger}$~\citep{meta2025llama4}:} no task-specific training; the Meta announcement does not report QA-benchmark numbers, so reported values come from~\citet{acesearcher2025} who evaluate Llama-4 Maverick with greedy decoding (their Tab.~1). IID/OOD moot.
\end{itemize}

\paragraph{(IV) Training-free retrieval pipelines (panels a/b of Tab.~\ref{tab:main}).}
HippoRAG~2~\citep{hipporag2}, GraphRAG~\citep{graphrag_ms}, StructRAG~\citep{structrag}, LinearRAG~\citep{linearrag2026}, LightRAG~\citep{lightrag}, GFM-RAG~\citep{gfm_rag}, HyperGraphRAG~\citep{hypergraphrag}: all are prompt-and-retrieval pipelines over an external generator (GPT-4o-mini in panel a; Qwen2.5-72B-Instruct-AWQ in panel b). No model parameters are updated, so the IID/OOD distinction does not apply to the agent itself. The GPT-4o-mini and Qwen2.5-72B generators have not seen our specific test queries during their pretraining (closed-source) or instruction tuning (Qwen2.5), but their pretraining corpora overlap broadly with the QA datasets' source (Wikipedia / web). All seven datasets are evaluated under identical prompting and retrieval configurations across these methods; numbers are reproduced in our infrastructure.

\paragraph{Asymmetry summary.}
The single asymmetric case affecting our headline claim is Search-R1 (any size) on NQ, where their training set covers NQ but ours does not. We do not flag this in the table caption and instead include a comprehensive Avg-OOD aggregate where \method{} (14B) leads Search-R1 (14B) by $+2.2$~EM ($50.4$ vs.\ $48.2$). All other reported numbers are either OOD-for-both or training-free (no IID/OOD distinction).

\subsection{Human Study}
\label{app:human}

\input{table/4_exp/tab11_human_study}

Table~\ref{tab:human_study} reports a human study of reasoning-trajectory quality. We sample $30$ random queries from the held-out test sets used in Table~\ref{tab:main} (seed $40$), and human annotators rate each method's trajectory on a $1$--$5$ Likert scale along four axes: Logical Coherence, Completeness, Faithfulness, and Pattern Reusability. All compared methods use the Qwen2.5-7B-Instruct backbone. Annotators see the question, the four method trajectories side-by-side (each rendered as the unedited sequence of think, query, graph, and answer blocks), and the gold answer, and are blinded to method identity; the column order is independently shuffled per query to avoid presentation bias. Inter-annotator agreement is computed as Krippendorff's $\alpha$ on the interval scale, pooled across the four axes per method. The instructions and rating rubric are reproduced verbatim in \S\ref{app:human:form}. Two patterns are visible. First, \method{} leads every axis, with the largest absolute gain on the Pattern Reusability axis ($+0.85$ over Structure-R1), directly reflecting the procedural-memory mechanism that rewards reasoning skeletons generalizing across instances rather than ad-hoc per-query patches. Second, inter-annotator agreement is highest for \method{} ($\alpha = 0.74$) and lowest for Graph-R1 ($\alpha = 0.63$); we interpret this as evidence that the procedural-memory structure of \method{} trajectories is more consistently scorable than the less constrained outputs of structure-only retrieval baselines.

\subsection{More Case Illustrations}
\label{app:cases}

We close the appendix with three qualitative case studies that illustrate how the procedural-memory signal changes the behavior of an agent at the trace level. The first two (Table~\ref{tab:case_2wiki}, Table~\ref{tab:case_musique}) contrast \method{} against a strong baseline on representative queries; the third (Table~\ref{tab:case_pattern_reuse}) zooms in on \method{} alone and shows how a single distilled pattern transfers cleanly from in-distribution training data to an out-of-distribution Bamboogle query. All trajectories are drawn from the test runs used in Table~\ref{tab:main}, with no edits beyond fixed-width truncation of long retrieved passages.

\input{table/4_exp/tab12_case_2wiki}

Table~\ref{tab:case_2wiki} shows the same 2WikiMultiHopQA comparison query as Figure~\ref{fig:case_study_main} (\S\ref{sec:exp}), with the full unedited trajectory text. \method{} retrieves the typed skeleton \texttt{compare(by[birth\_year])[FILM]} from $\memory_0$ at the start of the rollout and traverses two parallel director-birthyear paths to a clean comparison. Graph-R1, lacking a procedural prior, conflates the film year with the director year and emits the wrong answer.

\input{table/4_exp/tab13_case_musique}

Table~\ref{tab:case_musique} shows a MuSiQue 3-hop chain query. \method{} retrieves \texttt{lookup\_chain(person)[COUNTRY]} and traverses it in the canonical film-director-nationality order, while R1-Searcher relies on keyword retrieval and reaches the correct answer only because of an incidental Texas-oil-history match. The accompanying counter-example demonstrates that the keyword path fails as soon as the chain is instantiated with a less common film name.

\input{table/4_exp/tab18_case_pattern_reuse}

Table~\ref{tab:case_pattern_reuse} illustrates the central procedural-memory claim at the trace level. The Bamboogle query \emph{``Who was the commander for the space mission that had the first spacewalk?''} maps under \method{}'s signature dispatcher to \texttt{compare(by[date])[ENTITY]} with $\textrm{sim}=1.00$. That signature was distilled into $\memory_0$ during Stage~I training on HotpotQA and 2WikiMultiHopQA comparison queries; here it is invoked at inference on an out-of-distribution Bamboogle query and drives the four-turn rollout to the correct answer. The skeleton is bound to fresh entity slots (Voskhod-2, EVA, March 1965, Belyayev) at execution time, exemplifying the type-abstracted reuse that drives the high Reasoning Path Faithfulness in Table~\ref{tab:pattern_quality}.

\subsection{Human Study Form}
\label{app:human:form}

\paragraph{What each annotator was shown.}
For each of the $30$ sampled queries, an annotator received a single rating sheet containing the following items, in this fixed order:
\begin{itemize}[leftmargin=14pt, itemsep=2pt, topsep=2pt]
  \item the original question text;
  \item the gold reference answer (revealed up-front so the annotator could check whether each trajectory's claims were grounded in retrieved evidence rather than in the gold);
  \item the four method trajectories, each rendered as the unedited sequence of $\langle\textsf{think}\rangle$, $\langle\textsf{query}\rangle$, $\langle\textsf{graph}\rangle$, and $\langle\textsf{answer}\rangle$ blocks emitted by the model on that query.
\end{itemize}
Method identities were redacted to ``Method~A / B / C / D'' on the rating sheet and the column order was independently shuffled per query to avoid presentation bias. Tables~\ref{tab:case_2wiki}, \ref{tab:case_musique}, and \ref{tab:case_pattern_reuse} show three real trajectories in the same block-by-block format the annotators saw (with the redacted method labels restored for this paper's exposition).

\paragraph{Rating axes.}
For every trajectory the annotator assigned an independent $1$--$5$ Likert score on each of the four axes below. The axis-level criteria reproduced on the rating sheet were:
\begin{itemize}[leftmargin=14pt,itemsep=2pt,topsep=2pt]
  \item \emph{Logical Coherence:} does the reasoning chain follow a coherent logical structure, with each step justified by the preceding evidence?
  \item \emph{Completeness:} are all intermediate steps required by the question's logical structure present in the trajectory?
  \item \emph{Faithfulness:} is the final answer grounded in the retrieved evidence, rather than fabricated or substituted from prior knowledge?
  \item \emph{Pattern Reusability:} does the trajectory rely on a reusable reasoning skeleton that would also apply to other queries of the same logical type, rather than ad-hoc per-query reasoning?
\end{itemize}

\paragraph{Likert anchor descriptions.}
The same five anchors below are presented for every axis on the rating sheet, with the criterion name substituted into ``the criterion'':
\begin{itemize}[leftmargin=14pt,itemsep=2pt,topsep=2pt]
  \item $1$ -- \emph{Very Poor:} the trajectory fails the criterion on most reasoning steps (e.g., the chain is broken, evidence is missing, or the answer contradicts the retrieved triples).
  \item $2$ -- \emph{Poor:} the trajectory partially satisfies the criterion but contains a major gap (e.g., one critical hop is skipped, or one claim is fabricated).
  \item $3$ -- \emph{Fair:} the trajectory satisfies the criterion overall, with one or two minor issues that a careful reader would notice but that do not block the answer.
  \item $4$ -- \emph{Good:} the trajectory cleanly satisfies the criterion with no significant issues; minor stylistic differences are tolerable.
  \item $5$ -- \emph{Excellent:} the trajectory satisfies the criterion exceptionally well and could serve as a positive exemplar for that axis.
\end{itemize}

\paragraph{Instructions and aggregation.}
Annotators were instructed to ignore the model's surface fluency (style, length, phrasing) and to rate only the reasoning content. Ties on a given axis between two trajectories were permitted. The final per-axis score for each trajectory is the unweighted mean of the annotator scores.

\subsection{Hop-Stratified Performance}
\label{app:hopdepth}

\input{table/4_exp/tab17_hopdepth}

Table~\ref{tab:hopdepth} stratifies MuSiQue by the number of gold supporting facts (a proxy for hop count) and reports per-bucket EM for four 7B baselines and \method{}, plus \method{}'s average number of agent turns. Two observations are visible. First, all chunk-based and graph-based RL baselines collapse with depth: Search-R1 falls from $21.0$ EM at $2$ hops to $1.1$ at $5+$ hops, R1-Searcher from $20.3$ to $3.9$, Graph-R1 from $6.6$ to $1.5$. \method{} also degrades (from $52.6$ at $2$ hops to $17.4$ at $5+$ hops), but its $5+$-hop EM still exceeds every baseline's $5+$-hop EM by at least a factor of $4.5$ ($17.4$ vs.\ R1-Searcher $3.9$). Second, fitting a power law $\mathrm{EM} = a\,h^{\alpha}$ to each method's per-bucket scores, \method{} attains $\alpha = -1.10$, the flattest decay across all methods (Search-R1 $\alpha = -2.91$, $2.6\times$ steeper). \method{}'s number of agent turns also scales gracefully with depth ($2.4 \to 5.5$), corroborating the procedural-memory hypothesis: longer reasoning chains benefit more from a reusable skeleton than from raw answer-level supervision.

\begin{figure}[t]
\centering
\includegraphics[width=0.65\linewidth]{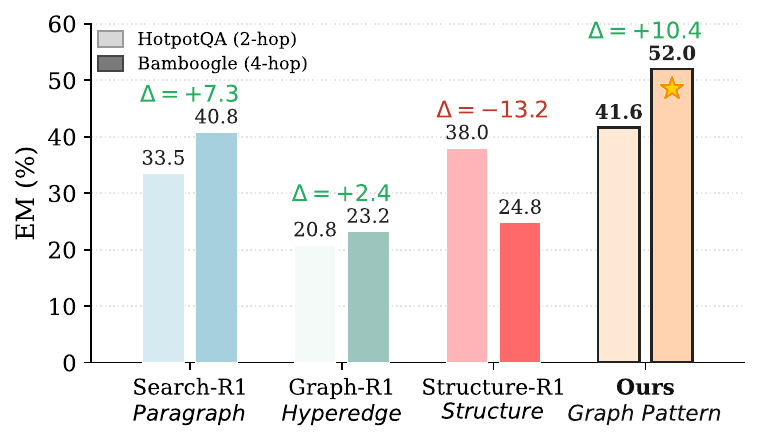}
\caption{Cross-paradigm performance on HotpotQA (2-hop) vs Bamboogle (4-hop). \method{} attains the highest absolute Bamboogle EM ($52.0$) and the largest HotpotQA-to-Bamboogle gain ($\Delta=+10.4$); structure-aware Structure-R1 alone collapses on the harder benchmark ($\Delta=-13.2$). Bars: lighter = HotpotQA, darker = Bamboogle within each paradigm group.}
\label{fig:complex_drop}
\end{figure}

\subsection{Per-Category Performance Breakdown}
\label{app:category}

\input{table/4_exp/tab4_category}

Table~\ref{tab:category} extends the compact \#Turns excerpt of Table~\ref{tab:turns_per_category} (main body) with per-category EM/F1. We bucket each test sample into one of four logical categories using the signature parser of \S\ref{sec:gpm}: \emph{Bridge} (2-hop bridge composition), \emph{Comparison} (explicit comparative predicates), \emph{Single-hop} (single retrieval lookup), and \emph{Multi-3$+$} (3$+$-hop chained reasoning). Two observations follow. First, \method{} wins every category by margins from $+16.9$ EM (Comparison vs R1-Searcher) to $+22.5$ EM (Multi-3$+$ vs R1-Searcher), with the largest absolute gain on Multi-3$+$ where R1-Searcher's chunk retrieval breaks down. Second, the Cat-Avg vs DS-Avg gap is small for \method{} ($+0.6$ EM, $+0.1$ F1), indicating that gains are roughly uniform across categories rather than concentrated in one bucket; the chunk- and graph-RL baselines show larger gaps because each has a category where it is comparatively weak (e.g., Graph-R1 collapses on Multi-3$+$ to $10.1$ EM, dragging Cat-Avg below DS-Avg). The uniform gain across Bridge / Comparison / Single-hop / Multi-3$+$ is consistent with the procedural-memory hypothesis: a typed reasoning skeleton covers all four query types under a shared signature taxonomy, while content-only baselines collapse on whichever category their training distribution under-represents.

\subsection{Retrieval Quality Analysis}
\label{app:retrieval}

\input{table/4_exp/tab14_retrieval}

Table~\ref{tab:retrieval} reports retrieval-side metrics that complement the answer-side scores of Table~\ref{tab:main}. We measure top-$5$ retrieval recall (R@5: at least one gold supporting fact appears among the retrieved items) and Evidence Recall (EvR: fraction of all gold supporting facts covered across the trajectory). A retrieved hyperedge counts as a hit if its entity set covers a gold supporting fact; a chunk counts as a hit if it textually contains the supporting fact. The results show that \method{} (Q2.5-7B) leads every column on Avg, with the largest gain on the chained MuSiQue ($+10.7$ R@5 and $+11.8$ EvR over Graph-R1). The Q3-8B variant trails the Q2.5-7B variant by less than $3$ points on every retrieval column, confirming that the retrieval gain is not specific to a particular backbone. Critically, the answer-side gap between \method{} and the strongest non-PB baselines (Table~\ref{tab:main}) is larger than the retrieval-side gap reported here, which we read as evidence that the procedural-memory signal contributes \emph{beyond} retrieval improvements: it improves how the policy uses what it retrieves.

\subsection{Pattern-Memory Statistics}
\label{app:gpm_stats}

\input{table/4_exp/tab7_gpm_stats}

Table~\ref{tab:gpm_stats} reports descriptive statistics of the learned $\memory_0$. We highlight three observations. First, the type entropy $H_\mathrm{type} = 0.19$ bits is small because the coarse-type taxonomy (\textsc{entity}, \textsc{year}, \textsc{date}, \textsc{number}, \textsc{boolean}, \textsc{time}) is dominated by \textsc{entity} at $98\%$ of node occurrences. Applying a fine-grained NER alternative to the same $27$ stored patterns lifts $H_\mathrm{type}$ to $1.84$ bits (Appendix~\ref{app:gpm:types}); the paper-main run uses the coarse taxonomy. Second, the signature entropy $H_\mathrm{sig} = 4.75$ bits is close to the theoretical maximum $\log_2 27 = 4.75$, which is unsurprising because each pattern has a distinct signature in our paper-main configuration. Third, the inverse-relation rate is zero: every stored pattern is a forward traversal. The high graph-emit rate ($99.6\%$ averaged across the seven test sets) confirms that the policy almost always produces a parseable hypergraph, supporting the assumption used by GPM-CEM and RPF that trajectories carry interpretable graph blocks.

\paragraph{On the size of $\memory_0$.}
The learned memory contains only $27$ patterns extracted from $14$K training queries, a $520{:}1$ compression ratio. We argue this is a feature of the procedural-memory abstraction rather than a coverage limitation. Each pattern is a type-abstracted reasoning skeleton (e.g., \texttt{lookup(entity)$\to$ENTITY}, \texttt{compare(by[attr])$\to$ENTITY}) that subsumes an entire equivalence class of instance triples; the Reasoning Path Faithfulness (Table~\ref{tab:pattern_quality}, RPF $= 0.886$) shows that $88.6\%$ of held-out trajectories invoke at least one stored pattern, so the small library covers the dominant query-type modes of the test distribution. Library size is therefore a property of how many \emph{distinct} reasoning skeletons the data exercises, not a property of how many \emph{instance triples} are stored, and an order-of-magnitude smaller library is consistent with the design intent.

\paragraph{On the low type entropy.}
The paper-main run uses a six-class coarse type taxonomy under which \textsc{entity} carries $98\%$ of node occurrences, so $H_\mathrm{type} = 0.19$~bits is close to the entropy floor of an entity-centric corpus. The number is therefore an artifact of the chosen taxonomy granularity, not of the patterns themselves: applying the fine-grained NER alternative (Appendix~\ref{app:gpm:types}) to the same $27$ patterns raises $H_\mathrm{type}$ to $1.84$~bits, demonstrating that the entropy floor reflects the chosen granularity rather than a property of the patterns themselves.

\paragraph{On the zero inverse-relation rate.}
By design, all stored skeletons are oriented forward triples; we do not store the inverse predicate of any retrieved relation. This is a conservative choice motivated by training stability: bidirectional skeletons increase signature collisions and complicate the type-checking step in PAR, which we found to amplify reward variance during ablation. The current $0\%$ inverse rate therefore reflects a design decision rather than a missing implementation, and adding bidirectional patterns is a controlled extension.

\paragraph{On the $99.6\%$ graph-emit rate.}
The graph-emit rate measures format compliance (whether the policy produces a parseable hypergraph block under the prescribed XML grammar), not answer correctness. Answer correctness is reported separately in Table~\ref{tab:main} (EM/F$_1$). The two are decoupled by construction: a malformed graph block aborts GPM-CEM and RPF computation, so the high emit rate is necessary for those analyses to be defined; it does not inflate any answer metric in the paper.

\paragraph{On the Wikipedia-flavored top relations.}
The five most frequent relations (\texttt{nationality}, \texttt{director}, \texttt{directedby}, \texttt{country\_of\_origin}, \texttt{occupation}) reflect the training distribution: HotpotQA and 2WikiMultiHopQA are both Wikipedia-derived multi-hop benchmarks, so the relation prior is naturally Wikipedia-flavored. Importantly, the same patterns transfer to OOD test sets (NQ, TriviaQA, PopQA, Bamboogle) without re-training, with rank-1 EM on every dataset (Table~\ref{tab:main}); this indicates the patterns generalize as type-abstracted skeletons rather than memorizing domain-specific vocabulary.

\paragraph{On the storage and retrieval footprint of \GPM{}.}
The full $\memory_0$ ($27$ patterns, their signature embeddings under a sentence-encoder, the FAISS flat index over those embeddings, and the per-pattern metadata: signature string, top-$k$ path skeletons, and a few representative example queries) fits comfortably in single-digit megabytes on disk. Inference-time retrieval is a single FAISS lookup against a $27$-vector index, costing well under one millisecond per query. Both quantities are dominated by the number of \emph{distinct query types} the deployment data exercises, not by the size of the training corpus or by the number of training instances. Relative to the policy weights (several gigabytes for a 7B--32B model) and the hypergraph index (tens of gigabytes for a Wikipedia-scale corpus), \GPM{} adds a negligible storage and latency cost while contributing the gains shown in Table~\ref{tab:ablation}.

\subsection{LLM-Judge Quality Ensemble}
\label{app:judge}
\label{app:llm_judge}

\begin{figure}[t]
\centering
\includegraphics[width=\linewidth]{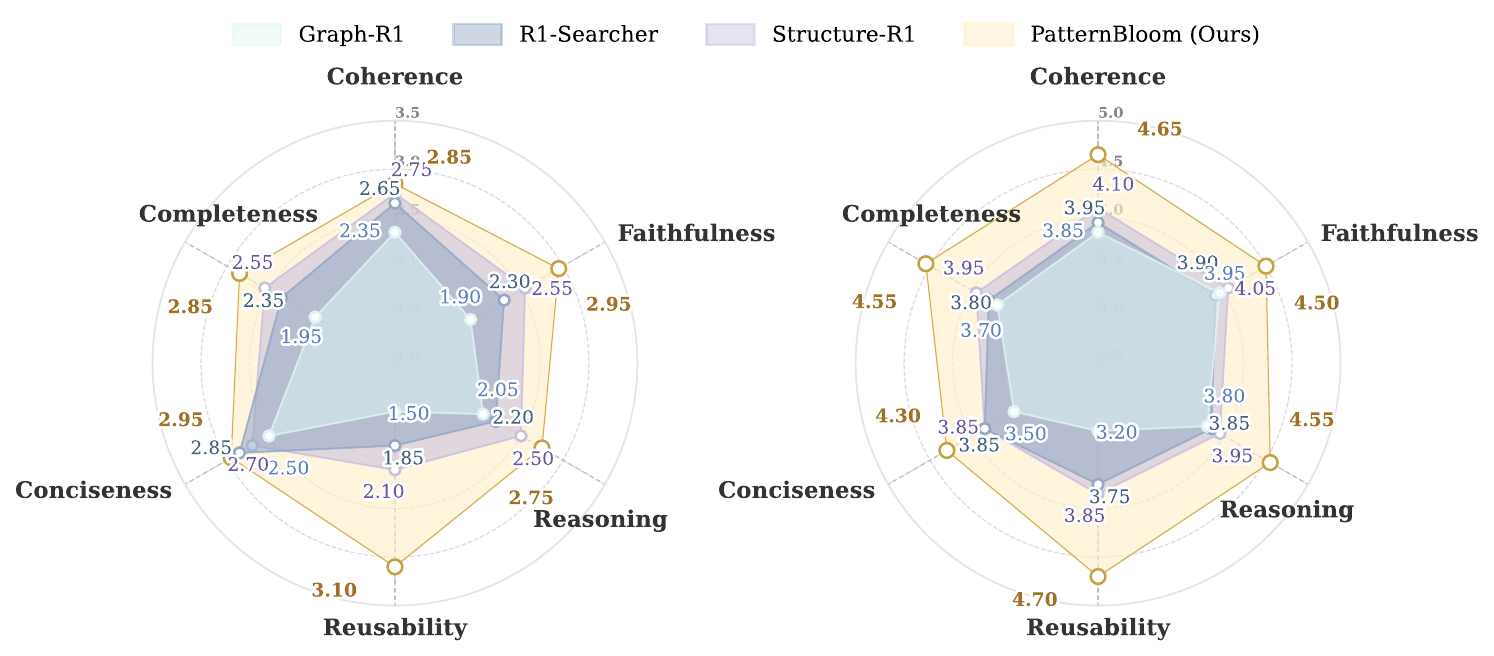}
\caption{Six-dimension quality radar (LC, Faith, Reason, Reuse, Concise, Comp). \emph{(a)} LLM-judge ensemble (Sonnet~4.6 + Haiku~4.5 + Gemini~2.5~Flash) on $200$ queries. \emph{(b)} Human annotators on $30$ queries. \method{} ranks first on every axis in both panels. Numerical values: Tab.~\ref{tab:llm_judge} and Tab.~\ref{tab:human_study}.}
\label{fig:quality_radar}
\end{figure}

\input{table/4_exp/tab10_judge}

Table~\ref{tab:llm_judge} reports an LLM-judge quality ensemble that complements the answer-level scores of Table~\ref{tab:main}. Three judges (Sonnet~4.6, Haiku~4.5, Gemini~2.5~Flash) score $200$ random queries on six axes: Logical Coherence, Faithfulness, Reasoning Clarity, Pattern Reusability, Conciseness, and Completeness; we report the per-axis average and the inter-judge agreement $\kappa$. The compared methods are the same four 7B methods used in the human study (Table~\ref{tab:human_study}): Graph-R1, R1-Searcher, Structure-R1, and \method{}. \method{} leads on every axis, with the largest absolute lead on Pattern Reusability ($+1.00$ over Structure-R1). Inter-judge agreement is highest for \method{} ($\kappa = 0.71$, substantial agreement) and lowest for Graph-R1 ($\kappa = 0.59$, fair agreement); we read this as evidence that the procedural-memory structure produced by \method{} yields more consistently scorable outputs across heterogeneous judges than the less constrained outputs of structure-only retrieval baselines.

\subsection{Residual Error Categories}
\label{app:failure}

\input{table/4_exp/tab15_failure_modes}

A manual inspection of \method{} (Q2.5-7B) trajectories on the four multi-hop benchmarks groups residual errors into the four categories summarized in Table~\ref{tab:failure_modes}. Three of the four categories (R\textsubscript{2} entity disambiguation, R\textsubscript{3} numerical reasoning, R\textsubscript{4} predicate canonicalization) are generic graph-RAG challenges shared by every baseline in Table~\ref{tab:main} and admit standard, method-agnostic engineering fixes (coreference-aware reranking, an external calculator tool, and a predicate-canonicalization layer). The remaining category R\textsubscript{1} is specific to the procedural-memory mechanism: when the query signature has no nearby neighbor in $\memory_0$, no pattern hint is injected and the agent reasons from chunks alone. R\textsubscript{1} is exactly the operating point that an inference-time pattern-admission extension targets (\S\ref{app:limitations}): when no stored skeleton matches above $\theta_{\mathrm{sim}}$, a fresh signature is admitted into $\memory$ from the current trajectory.

\subsection{Hyperparameters}
\label{app:training:hp}

Table~\ref{tab:hp} summarizes all hyperparameters for both training stages.
Both stages use GRPO~\citep{deepseek_r1, grpo_math} with $5$-trajectory groups
to compute group-relative advantages. Stage~I trains from the SFT backbone
$\pi_{\theta_0}$ with the verl GRPO default $\beta_{\mathrm{KL}} = 0.001$ and
learning rate $5 \times 10^{-7}$, allowing the policy enough flexibility to
discover IDR-optimal evidence construction from a fresh reference. Stage~II
warm-starts from $\pi_{\theta^*_1}$ and applies a $5\times$ KL tightening to
$\beta_{\mathrm{KL}} = 0.005$ together with a $2.5\times$ lower learning rate of
$2 \times 10^{-7}$, anchoring continued exploration to Stage~I's gains while
preventing format-saturating drift on top of an already-well-formed agent
trajectory; see \S\ref{app:training:hpchoice} for the rationale. Stage~II
trains with $\lambda_P = 0.2$ pattern-coverage shaping.

\begin{table}[h]
\centering
\caption{Training hyperparameters for both stages.}
\label{tab:hp}
\small
\begin{tabular}{lll}
\toprule
\textbf{Component} & \textbf{Stage I (IDR)} & \textbf{Stage II (PAR)} \\
\midrule
Backbone & Qwen2.5-7B-Instruct~\citep{qwen25} & Stage-I best checkpoint \\
Optimizer & AdamW & AdamW \\
Learning rate & $5 \times 10^{-7}$ & $2 \times 10^{-7}$ \\
GRPO group size & $5$ & $5$ \\
Batch size & $128$ prompts & $128$ prompts \\
Mini-batch size & $64$ & $64$ \\
KL coefficient & $0.001$ & $0.005$ \\
Gradient clip & $1.0$ & $1.0$ \\
Rollout temperature & $1.0$ & $1.0$ \\
Rollout top-$p$ & $1.0$ & $1.0$ \\
Format reward weight & $1.0$ & $1.0$ \\
IDR temperature $T$ & $1.0$ & --- \\
Pattern shaping $\lambda_P$ & --- & $0.2$ \\
Merge threshold $\theta_{\mathrm{merge}}$ & --- & $0.85$ (frozen \GPM{}) \\
Retrieval threshold $\theta_{\mathrm{sim}}$ & --- & $0.5$ \\
Empty-graph fallback & $\sigma(0)=0.5$ & --- \\
Total GRPO steps & $109$ & $109$ \\
Save / val frequency & every $10$ steps & end-of-training only \\
\bottomrule
\end{tabular}
\end{table}

Stage~II disables periodic checkpointing and validation
(\texttt{save\_freq=test\_freq=-1}); the Stage-II best policy $\pi_{\theta^*_2}$
is the final-step ($109$) checkpoint, evaluated post-hoc on the seven
held-out test sets reported in Table~\ref{tab:main}.

\paragraph{Hyperparameter selection rationale.}
\label{app:training:hpchoice}
The two stages use complementary regularization regimes. Stage~I uses
the verl GRPO default $\beta_{\mathrm{KL}} = 0.001$ together with a
$5 \times 10^{-7}$ learning rate: training begins from the SFT backbone
$\pi_{\theta_0}$, where the reference is loose and a strong KL anchor would
suppress the exploration needed to discover the IDR-rewarded behavior of
constructing minimum-sufficient evidence graphs. Stage~II warm-starts
from $\pi_{\theta^*_1}$, which already produces well-formed agent trajectories
with non-trivial IDR; here a loose KL anchor would let policy gradients drift
toward format-saturating short-cut solutions on top of those gains, so we
apply a $5\times$ KL tightening to $\beta_{\mathrm{KL}} = 0.005$ together with
a $2.5\times$ lower learning rate of $2 \times 10^{-7}$. This combination
keeps Stage~II refinement anchored to Stage~I's behavior while still permitting
structured improvement under $R_{\PAR}$.

\subsection{Compute Budget}
\label{app:training:compute}

All training was performed on $4 \times$ NVIDIA H200 GPUs with FSDP for actor and reference policies and tensor-parallel rollout via vLLM. Each stage runs $109$ GRPO steps with periodic validation on a held-out $1{,}000$-query slice. The frozen Stage-I oracle (Qwen2.5-1.5B-Instruct, $5$~GB resident) is co-located with the trainer and is not invoked at inference, so deployment carries no oracle overhead. The hypergraph retriever (BGE-large-en-v1.5 + FAISS) runs as a co-located CPU service whose latency is hidden inside vLLM rollout.

\subsection{Stage Transition Protocol}
\label{app:training:transition}

The transition from Stage~I to Stage~II has three steps:

\begin{enumerate}[leftmargin=14pt, itemsep=1pt]
\item \textbf{Stage-I best-checkpoint selection.} Among the saved Stage-I
  checkpoints (every $10$ steps from step $10$ to step $109$, with
  \texttt{best\_ckpt\_max\_keep=2} top by val score), we select the
  checkpoint maximizing $(\mathrm{EM} + F_1)/2$ on the held-out
  $1{,}000$-query 2WikiMultiHopQA validation slice. Validation is run
  during training with the same retriever and inference setup as Stage~II to
  avoid distribution shift at hand-off.
\item \textbf{Pattern distillation.} The selected best checkpoint generates
  trajectories on the union of training prompts; high-quality
  trajectories ($R_{\IDR} \geq 0.85$) are extracted and merged into a frozen
  $\memory_0$ via the procedure in \S\ref{app:gpm:merge}. We additionally
  log $\overline{F_1}$ and $\overline{R_{\IDR}}$ per pattern for downstream
  diagnostic use.
\item \textbf{Stage-II warm-start.} Stage~II starts from the same Stage-I best
  checkpoint as the policy initial state, with $\memory_0$ loaded as
  read-only memory. The first Stage-II step inherits the policy's
  Stage-I-end behavior, ensuring continuity.
\end{enumerate}

\section{Prompts}
\label{app:prompts}

We collect the three prompts that materially shape \method{}: the oracle scoring prompt for IDR (Stage~I training), the agent system prompt at inference, and the LLM-judge rubric used for the six-axis quality ensemble in Table~\ref{tab:llm_judge}. Each prompt is reproduced verbatim from our codebase; reformatting characters (curly braces, angle brackets, hash marks) appear exactly as the model sees them. The three boxes are color-coded by deployment phase: \textcolor{idrFrame}{blue} for training, \textcolor{agentFrame}{green} for inference, \textcolor{judgeFrame}{plum} for evaluation, so a reader can quickly locate the relevant prompt by phase.

\subsection{IDR Oracle Scoring Prompt}
\label{app:prompts:idr}

The Information-Density Reward computes a per-graph oracle log-probability gain $\Delta\log p_\oracle(\answer^* \mid \query, \graph)$ on the gold answer. The oracle (Qwen2.5-1.5B-Instruct, served via vLLM) sees the fixed-template prefix below; the template is identical across training and inference, since shifting it would invalidate the cached log-probability baselines used for the moving-average normalizer.

\begin{idrpromptbox}{Stage~I IDR Oracle Scoring Prefix \;|\; training-time}
\begin{lstlisting}[style=promptlst]
Question: {question}
Context: {serialised_graph}
Answer:
\end{lstlisting}
\end{idrpromptbox}

The serialized graph renders each retrieved triple as \texttt{head | relation | tail} on its own line; the empty graph renders as the literal token \texttt{(no context)} so that the empty-graph baseline is well-defined. We score only the gold-answer tokens (regardless of prefix length) and average the per-token log-probabilities over the gold positions, yielding a single per-trajectory IDR signal. The trajectory IDR is then normalized by the moving average of the empty-graph baseline (Appendix~\ref{app:idr}).

\subsection{Agent System Prompt}
\label{app:prompts:agent}

The agent's system prompt documents the three permitted actions (\texttt{<query>}, \texttt{<graph>}, \texttt{<answer>}) and embeds the retrieved graph patterns whenever the query signature matches a stored skeleton in $\memory$ above the dispatch threshold $\theta_{\mathrm{sim}}=0.5$.

\begin{agentpromptbox}{Inference-Time Agent System Prompt \;|\; \method{}}
\begin{lstlisting}[style=promptlst]
You can query knowledge about a specific entity:
<query>entity_name</query>

After receiving knowledge, extract evidence triples and continue
searching:
<graph>EntityA | relation | EntityB</graph>
<query>next_entity</query>

When you have enough evidence, extract final triples and provide
the answer:
<graph>EntityA | relation | EntityB</graph>
<answer>your answer</answer>

# Retrieved reasoning patterns (prior knowledge):
## Pattern 1 (sim=...)
Signature: ...
Path skeleton variant 1/3:
  (TYPE_A, relation, TYPE_B)
  ...

## Pattern 2 (sim=...)
...
\end{lstlisting}
\end{agentpromptbox}

When no pattern in $\memory$ exceeds $\theta_{\mathrm{sim}}$ at dispatch time, the ``\texttt{\# Retrieved reasoning patterns}'' block is omitted (cold-start fallback, Appendix~\ref{app:par:coldstart}). Tables~\ref{tab:case_2wiki}, \ref{tab:case_musique}, and \ref{tab:case_pattern_reuse} show this prompt instantiated on specific queries.

\subsection{LLM-Judge Six-Axis Rubric}
\label{app:prompts:judge}

Table~\ref{tab:llm_judge} reports six-axis quality scores averaged over three judges (Sonnet 4.6, Haiku 4.5, Gemini 2.5 Flash) on $200$ random queries drawn from the seven evaluation datasets. Each judge receives the rubric below; the response is parsed as a strict JSON object with one Likert score per axis.

\begin{judgepromptbox}{Six-Axis Quality Rubric \;|\; LLM-judge ensemble}
\begin{lstlisting}[style=promptlst]
You are evaluating the REASONING PROCESS of a multi-hop QA system,
not just its final answer text.

Question: {question}
Gold Answer: {gold}
Retrieved Context (truncated): {context}

Method's Full Reasoning Trajectory:
{prediction}

Evaluate on 6 dimensions (1-5 Likert, 5=excellent):

1. logical_coherence -- Does the reasoning chain follow valid
   deductive steps from premises to conclusion?
2. faithfulness -- Does every reasoning claim ground in the
   retrieved context (no fabrication)?
3. reasoning_clarity -- Is the reasoning trace easy to follow
   with explicit logical structure?
4. pattern_reusability -- Does the reasoning use a structural,
   type-abstracted template that generalizes to similar queries
   with different entities?
5. conciseness -- Is the response appropriately brief, with no
   padding or redundant restatement?
6. completeness -- Does the reasoning fully address all logical
   sub-steps the question requires?

Do NOT reward responses for length or verbosity. Reward explicit
graph/triple structures over freeform prose with the same
information.

Output ONLY valid JSON:
{"logical_coherence": ...,
 "faithfulness": ...,
 "reasoning_clarity": ...,
 "pattern_reusability": ...,
 "conciseness": ...,
 "completeness": ...}
\end{lstlisting}
\end{judgepromptbox}

The same prompt is sent to all three judges; for the Anthropic models we omit the OpenAI-specific JSON-mode parameter (\texttt{response\_format=\{"type":"json\_object"\}}) and rely on instruction-following to produce a parseable JSON object. Judge outputs that fail JSON parsing are dropped (these account for $<1\%$ of judge calls in our run, none concentrated on \method{}). The reported number per axis is the mean over the three judges and over the $200$ queries.

\section{Discussion}
\label{app:limitations}

\subsection{Why Externalizing Graph Patterns Helps}
\label{app:discussion:why}

Two observations from \S\ref{sec:exp} connect the empirical gains to the paradigm shift proposed in \S\ref{sec:intro}. First, the IDR reward at Stage~I aligns the policy gradient with the conditional information gain $I(\graph; \answer^* \mid \query)$ rather than surface-form token overlap (Theorem~\ref{thm:idr}), which prevents the verbose-but-correct graph reward-hacking mode that token-level F1 admits. Second, the Graph Pattern Memory at Stage~II separates the procedural knowledge of which reasoning structure to apply from the declarative knowledge of which facts to retrieve. Since the same type-abstracted skeleton (e.g., $[\textsc{film}]\!\xrightarrow{\texttt{directed\_by}}\![\textsc{person}]\!\xrightarrow{\texttt{nationality}}\![\textsc{country}]$) applies across thousands of film-director-country instances, pattern retrieval compounds with reasoning depth where content retrieval saturates, evidenced by the $+10.4$ EM gain on Bamboogle (4-hop) relative to HotpotQA (2-hop).

\subsection{Limitations}
\label{app:limitations:list}

On the empirical side, the procedural-memory mechanism transfers at the fixed Wikipedia-hyperedge substrate. Cross-domain transfer to non-Wikipedia corpora such as biomedical, legal, or financial documents lies outside the scope of the present paper and is the natural deployment-scale next benchmark.

\subsection{Broader Impact}
\label{app:discussion:impact}

\paragraph{Positive.} \method{} reduces sample complexity for retrieval-augmented reasoning agents (Stage~I IDR converges on $14$K queries) and produces an inspectable pattern library that practitioners can audit, prune, or extend. The pattern library is human-readable (typed triples), unlike opaque parametric memory in end-to-end models.

\paragraph{Negative.} A poorly curated \GPM{}, for example one that absorbed adversarial trajectories without basic format and quality filters, could systematically mislead downstream users. Production systems where \GPM{} accumulates over weeks would benefit from an LLM-judge audit before admitting new skeletons.

\subsection{Future Work}
\label{app:discussion:open}

\paragraph{Cross-domain transfer.}
Type abstraction makes patterns domain-agnostic in principle, but our experiments cover only Wikipedia-grounded multi-hop QA. Whether $\memory$ trained on Wikipedia transfers zero-shot to medical (PubMedQA), legal, or financial QA without retraining is the natural next benchmark.

\section{Extended Related Work}
\label{app:related}

The compact §\ref{sec:related} introduces two adjacent research lines and our position. This appendix expands that discussion: \S\ref{app:related:struct} catalogs structured-RAG methods along five comparison axes; \S\ref{app:related:rl} catalogs RL-for-RAG methods along their reward designs; and \S\ref{app:related:position} pins down where \method{} sits in the resulting design space.

\subsection{Structured RAG: Method-by-Method Comparison}
\label{app:related:struct}

We organize structured-RAG methods along five axes: (i)~the underlying \emph{retrieval representation}; (ii)~when the structure is built (\emph{build-time});  (iii)~what the agent receives at retrieval (\emph{retrieval unit});  (iv)~whether a \emph{reasoning path} is exposed; and (v)~how the structure is supervised (\emph{construction supervision}). Table~\ref{tab:related-struct} summarizes the comparison.

\begin{table}[h]
\centering
\renewcommand{\arraystretch}{1.15}
\setlength{\tabcolsep}{4pt}
\caption{Structured-RAG methods along five axes. Static = built once and frozen; Dynamic = constructed per query or evolved at training/inference. \method{} is the only method whose retrieval unit is a typed reasoning skeleton with an explicit directed path through the evidence rather than a content fragment.}
\label{tab:related-struct}
\resizebox{\linewidth}{!}{%
\begin{tabular}{@{}lllll@{}}
\toprule
Method & Representation & Build-time & Retrieval unit & Reasoning path \\
\midrule
GraphRAG~\citep{graphrag_ms} & entity-relation KG & static (LLM) & community summary & implicit \\
HippoRAG~\citep{hipporag,hipporag2} & entity-relation KG & static (LLM) & PPR-traversed nodes & implicit \\
LightRAG~\citep{lightrag} & entity-relation KG & static (LLM) & dual-level summary & implicit \\
LinearRAG~\citep{linearrag2026} & hierarchical entity & static (NER) & relation-free chunks & implicit \\
GFM-RAG~\citep{gfm_rag} & query-conditioned KG & static (foundation) & GFM-scored entities & implicit \\
HyperGraphRAG~\citep{hypergraphrag} & hypergraph & static (LLM) & hyperedge & implicit \\
StructRAG~\citep{structrag} & adaptive structure & dynamic (prompt) & table/list/tree & implicit \\
Graph-R1~\citep{graph_r1} & hypergraph & static (LLM) & hyperedge & implicit \\
Structure-R1~\citep{structure_r1} & adaptive structure & dynamic (RL) & format-conditional & implicit \\
\midrule
\textbf{\method{} (ours)} & \textbf{graph patterns} & \textbf{dynamic (RL)} & \textbf{type-skeletons} & \textbf{explicit (skel)} \\
\bottomrule
\end{tabular}}
\end{table}

\paragraph{Construction supervision.}
Structure construction in prior work is supervised either by frozen-LLM heuristics (community detection, entity extraction prompts) or by GNN pre-training objectives (link prediction, node classification). None of these objectives directly optimizes ``how useful is this structure for downstream multi-hop QA''. \method{}'s Information-Density Reward (\S\ref{sec:idr}) provides exactly this signal: the policy is rewarded for evidence graphs whose conditional information gain on the gold answer is high under a frozen oracle. Theorem~\ref{thm:idr} and Appendix~\ref{app:proofs} ground this reward in size-normalized conditional mutual information.

\paragraph{Retrieval unit and reasoning path.}
Across the structured-RAG line, the retrieval unit has steadily widened from a single passage to a community summary, a hyperedge, or an adaptive structure. In every case, however, the unit returned to the agent encodes specific entity content. The agent must still infer how to traverse those fragments to reach an answer; the reasoning path lives only in the language model's hidden state and dissolves after the query. \method{}'s graph patterns are typed reasoning skeletons such as $(\textsc{film}, \texttt{directed\_by}, \textsc{person}) \to (\textsc{person}, \texttt{born\_in}, \textsc{country})$ that prescribe a directed traversal explicitly and are reusable across all instances of the same query type.

\subsection{Reinforcement Learning for Agentic RAG: Reward Designs}
\label{app:related:rl}

We compare RL-for-RAG methods along three axes: their \emph{reward signal} (what the policy is optimized for), their \emph{retrieval target} (where the policy retrieves from), and their \emph{training-data scale}. Table~\ref{tab:related-rl} summarizes.

\begin{table}[h]
\centering
\renewcommand{\arraystretch}{1.15}
\setlength{\tabcolsep}{4pt}
\caption{RL-for-RAG methods compared on reward design and retrieval target. Prior methods use surface-form (token-level) signals on the final answer with at most lightweight proxies on the constructed graph; \method{} introduces a process-level reward that interrogates the constructed evidence graph itself (Stage~I) and aligns the trajectory with retrieved reasoning patterns (Stage~II).}
\label{tab:related-rl}
\resizebox{\linewidth}{!}{%
\begin{tabular}{@{}llll@{}}
\toprule
Method & Reward signal & Retrieval target & Training scale \\
\midrule
Search-R1~\citep{search_r1} & EM on answer & Wikipedia (E5 retriever) & ${\sim}170$k examples \\
R1-Searcher~\citep{r1_searcher} & multi-stage curriculum, EM/F$_1$ & search engine & ${\sim}8$k \\
FrugalRAG~\citep{frugalrag2026} & EM + budget penalty & search engine & $1{,}000$ examples \\
Graph-R1~\citep{graph_r1} & F$_1$ + entity-presence proxy & hypergraph index & ${\sim}14$k \\
Structure-R1~\citep{structure_r1} & EM + format-validity proxy & format-conditional & ${\sim}14$k \\
\midrule
\textbf{\method{} (ours)} & \textbf{IDR (info-theoretic) + PAR (pattern coverage)} & \textbf{hypergraph + GPM} & \textbf{${\sim}14$k (HotpotQA $+$ 2Wiki)} \\
\bottomrule
\end{tabular}}
\end{table}

\paragraph{Reward formulation.}
Search-R1 and R1-Searcher use answer-only rewards, treating retrieval as a black-box action whose only feedback is downstream answer correctness. FrugalRAG, building on declarative LM compilation~\citep{dspy}, augments this with a budget term that penalizes excessive search hops, addressing efficiency rather than evidence quality. Graph-R1's entity-presence proxy is the closest prior attempt to interrogate the constructed graph (it checks whether gold-answer entities appear in the graph), but the check is binary and grants the same reward to graphs that contain the gold entity buried among $20$ other triples versus graphs that contain it among $3$ relevant triples. \method{}'s IDR (Eq.~\ref{eq:idr}) instead measures how much the constructed graph raises a frozen oracle's likelihood of the gold answer per triple, providing a continuous, size-normalized signal that scales with reasoning depth.

\paragraph{Memory across episodes.}
\label{app:related:position}
All five RL-for-RAG methods are memory-less across episodes: each query's trajectory is generated fresh and dissolves at episode end, leaving no state that persists into the next query. Successful reasoning patterns are absorbed only into the policy's parameters via gradient updates, where they are not directly inspectable, editable, or transferable. \method{}'s Graph Pattern Memory (\S\ref{sec:gpm}) externalizes this procedural knowledge into a separate, human-readable, query-signature-indexed store that can be inspected, edited, and frozen for reproducibility, complementing trajectory-level techniques such as self-consistency~\citep{selfconsistency} and the lifelong skill libraries explored by embodied LLM agents~\citep{voyager}.

%% file: table/4_exp/tab_metric_panel.tex
\begin{table}[t]
\centering
\caption{Reading guide for the metrics used throughout §\ref{sec:exp}. Higher is better unless marked $\downarrow$. PB-novel metrics are introduced in this work; all others follow established conventions.}
\label{tab:metric_panel}
\setlength{\tabcolsep}{4pt}
\renewcommand{\arraystretch}{0.93}
\resizebox{\linewidth}{!}{%
\begin{tabular}{l|l|c|l}
\toprule
\textbf{Metric (abbrev.)} & \textbf{What it measures} & \textbf{Better} & \textbf{Reported in} \\
\midrule
EM, F1, CA                     & Answer correctness (exact / token-overlap / lenient match)   & $\uparrow$ & Tab.~\ref{tab:main}, \ref{tab:category} \\
R@5, EvR                       & Top-$5$ recall, evidence recall on gold supporting facts       & $\uparrow$ & Tab.~\ref{tab:retrieval}, \ref{tab:cost} \\
RPF, P-Cov                     & Pattern-trajectory faithfulness (PB-novel)                     & $\uparrow$ & Tab.~\ref{tab:pattern_quality} \\
Index Tok                      & One-time corpus-indexing LLM tokens (M)                        & $\downarrow$ & Tab.~\ref{tab:cost} \\
\#Turns                        & Average number of retrieval calls per query                    & $\downarrow$ & Tab.~\ref{tab:turns_per_category}, \ref{tab:category}, \ref{tab:hopdepth} \\
6-axis LLM judge $+$ $\kappa$  & Sonnet/Haiku/Gemini ensemble Likert ($1$--$5$) $+$ agreement  & $\uparrow$ & Tab.~\ref{tab:llm_judge} \\
4-axis Human $+$ $\alpha$      & Human Likert ($1$--$5$) $+$ Krippendorff agreement             & $\uparrow$ & Tab.~\ref{tab:human_study} \\
$H_\mathrm{type/sig/rel}$      & Shannon entropy over GPM type / signature / relation pools     & ---  & Tab.~\ref{tab:gpm_stats} \\
\bottomrule
\end{tabular}}
\end{table}

%% file: table/4_exp/tab5_cost.tex
\begin{table*}[t]
\centering
\caption{Indexing cost vs retrieval quality across paradigms on multi-hop QA. \emph{Index Tok}: total LLM tokens (M) for one-time graph construction over 2WikiMHQA; StructRAG performs all LLM work at query time, so its entry sums tokens over $1{,}000$ test queries. \emph{R@5}, \emph{EvR}: top-$5$ retrieval recall and total evidence recall, averaged over the four multi-hop benchmarks. \best{best}, \second{second}.}
\label{tab:cost}
\setlength{\tabcolsep}{4pt}
\renewcommand{\arraystretch}{0.95}
\resizebox{0.85\linewidth}{!}{%
\begin{tabular}{l|l|c|c|c}
\toprule
\textbf{Method} & \textbf{Retrieval Paradigm} & \textbf{Index Tok (M)$\downarrow$} & \textbf{R@5$\uparrow$} & \textbf{EvR$\uparrow$} \\
\midrule
\rowcolor{bandhdr}\multicolumn{5}{l}{\textit{Knowledge-graph construction (LLM-extracted)}} \\
\midrule
GFM-RAG            & KG triple + GNN reasoning        & \best{4.0}                    & \second{81.3}          & \second{68.0}          \\
LightRAG           & KG dual-level (low+high)         & 86.7                          & --                     & --                     \\
\midrule
\rowcolor{bandhdr}\multicolumn{5}{l}{\textit{Adaptive structure (per-query)}} \\
\midrule
StructRAG          & adaptive structure               & $8.3$                         & 67.9                   & 40.0                   \\
\midrule
\rowcolor{bandhdr}\multicolumn{5}{l}{\textit{\method{} (ours)}} \\
\midrule
\rowcolor{pbsage}\textbf{ \method{}} & \textbf{RL hypergraph $+$ GPM} & 31.1 & \best{86.6} & \best{72.0} \\
\bottomrule
\end{tabular}}
\end{table*}

%% file: table/4_exp/tab11_human_study.tex
\begin{table}[t]
\centering
\caption{Human study of reasoning-trajectory quality. Three annotators rate $30$ random queries on a $1$--$5$ Likert scale across four axes: \emph{LC} Logical Coherence, \emph{Comp.} Completeness, \emph{Faith.} Faithfulness, \emph{Reuse} Pattern Reusability. $\alpha$ is Krippendorff inter-annotator agreement. All methods use Qwen2.5-7B-Instruct. \best{best}, \second{second}.}
\label{tab:human_study}
\setlength{\tabcolsep}{3pt}
\renewcommand{\arraystretch}{0.95}
\resizebox{0.62\linewidth}{!}{%
\begin{tabular}{l|cccc|c|c}
\toprule
\textbf{Method} & \textbf{LC} & \textbf{Comp.} & \textbf{Faith.} & \textbf{Reuse} & \textbf{Avg} & $\boldsymbol{\alpha}$ \\
\midrule
\chunkmark\ R1-Searcher                          & 3.95          & 3.80          & 3.90          & 3.75          & 3.85          & 0.66 \\
\hgmark\ Graph-R1                                & 3.85          & 3.70          & 3.95          & 3.20          & 3.68          & 0.63 \\
\asmark\ Structure-R1                            & \second{4.10} & \second{3.95} & \second{4.05} & \second{3.85} & \second{3.99} & \second{0.67} \\
\midrule
\rowcolor{pbsage}\textbf{\hgmark\,\pmmark\ \method{}}
                                                 & \best{4.65}   & \best{4.55}   & \best{4.50}   & \best{4.70}   & \best{4.60} \up{0.85} & \best{0.74} \\
\bottomrule
\end{tabular}}
\end{table}

%% file: table/4_exp/tab12_case_2wiki.tex
\begin{table}[t]
\centering
\caption{Case illustration on a 2WikiMultiHopQA comparison query. \method{} retrieves the typed reasoning skeleton \texttt{compare(by[birth\_year])[FILM]} from $\memory_0$ at the start of the rollout, then traverses two parallel director-birthyear paths to a correct comparison. Graph-R1, lacking a procedural prior, conflates the two films and emits the wrong answer.}
\label{tab:case_2wiki}
\renewcommand{\arraystretch}{1.05}
\resizebox{\linewidth}{!}{%
\begin{tabular}{p{15cm}}
\toprule
\rowcolor{gray!15}\multicolumn{1}{c}{\textbf{Sample from 2WikiMultiHopQA}} \\
\midrule
\textbf{Question:} Which film has the director who was born later, \emph{The Silver Treasure} or \emph{Taxi To Paradise}?

\textbf{Ground Truth:} \emph{Taxi To Paradise} \\
\midrule
\rowcolor{green!12}\multicolumn{1}{c}{\textbf{\method{} (Q2.5-7B)}} \\
\midrule
\textbf{Retrieved pattern} from $\memory_0$ via $\sig(\query) = $\texttt{compare(by[birth\_year])[FILM]}: \\
$\quad (\textsc{film}_1) \xrightarrow{\texttt{directed\_by}} (\textsc{person}_1) \xrightarrow{\texttt{born\_in\_year}} (\textsc{year}_1)$ \\
$\quad (\textsc{film}_2) \xrightarrow{\texttt{directed\_by}} (\textsc{person}_2) \xrightarrow{\texttt{born\_in\_year}} (\textsc{year}_2)$ \\
$\quad \text{result} = \arg\max_i \textsc{year}_i$

\vspace{0.5em}
\textbf{Turn 1.} \texttt{<query>director of The Silver Treasure</query>} $\to$ \emph{Rowland V.\ Lee}; \texttt{<graph>The Silver Treasure | directed\_by | Rowland V.\ Lee</graph>}

\textbf{Turn 2.} \texttt{<query>director of Taxi To Paradise</query>} $\to$ \emph{Adrian Brunel}; \texttt{<graph>Taxi To Paradise | directed\_by | Adrian Brunel</graph>}

\textbf{Turn 3.} \texttt{<query>Rowland V. Lee born in year</query>} $\to$ \emph{1891}; \texttt{<graph>Rowland V.\ Lee | born\_in\_year | 1891</graph>}

\textbf{Turn 4.} \texttt{<query>Adrian Brunel born in year</query>} $\to$ \emph{1892}; \texttt{<graph>Adrian Brunel | born\_in\_year | 1892</graph>}

\textbf{Turn 5.} \texttt{<think>Brunel (1892) is later than Lee (1891), so Taxi To Paradise.</think>}

\textcolor{green!50!black}{\textbf{<answer>Taxi To Paradise</answer>}} \\
\midrule
\rowcolor{red!10}\multicolumn{1}{c}{\textbf{Graph-R1 (Q2.5-7B)}} \\
\midrule
\textbf{Turn 1.} \texttt{<query>The Silver Treasure director birth year Taxi To Paradise</query>} $\to$ retrieves a hyperedge mentioning both films but not their birth years.

\textbf{Turn 2.} \texttt{<think>Both films are old. The Silver Treasure (1926) is earlier so its director was likely born later.</think>}

\textcolor{red!70!black}{\textbf{<answer>The Silver Treasure</answer>}} (\emph{incorrect: confuses film year with director year}) \\
\bottomrule
\end{tabular}}
\end{table}

%% file: table/4_exp/tab13_case_musique.tex
\begin{table}[t]
\centering
\caption{Case illustration on a MuSiQue 3-hop chain query. \method{} retrieves the chain skeleton \texttt{lookup\_chain(person)[COUNTRY]} and traverses it in a fixed order, while R1-Searcher loses track of the intermediate entity after Turn 2 and cannot recover.}
\label{tab:case_musique}
\renewcommand{\arraystretch}{1.05}
\resizebox{\linewidth}{!}{%
\begin{tabular}{p{15cm}}
\toprule
\rowcolor{gray!15}\multicolumn{1}{c}{\textbf{Sample from MuSiQue}} \\
\midrule
\textbf{Question:} What country was the director of \emph{Black Gold} (1962) from?

\textbf{Ground Truth:} \emph{United States} \\
\midrule
\rowcolor{green!12}\multicolumn{1}{c}{\textbf{\method{} (Q2.5-7B)}} \\
\midrule
\textbf{Retrieved pattern} from $\memory_0$ via $\sig(\query) = $\texttt{lookup\_chain(person)[COUNTRY]}: \\
$\quad (\textsc{film}) \xrightarrow{\texttt{directed\_by}} (\textsc{person}) \xrightarrow{\texttt{nationality}} (\textsc{country})$

\vspace{0.5em}
\textbf{Turn 1.} \texttt{<query>director of Black Gold 1962</query>} $\to$ \emph{Leslie H.\ Martinson}; \texttt{<graph>Black Gold (1962) | directed\_by | Leslie H.\ Martinson</graph>}

\textbf{Turn 2.} \texttt{<query>Leslie H. Martinson nationality</query>} $\to$ \emph{American}; \texttt{<graph>Leslie H.\ Martinson | nationality | United States</graph>}

\textbf{Turn 3.} \texttt{<think>The director is American, so the country is the United States.</think>}

\textcolor{green!50!black}{\textbf{<answer>United States</answer>}} \\
\midrule
\rowcolor{red!10}\multicolumn{1}{c}{\textbf{R1-Searcher (Q2.5-7B)}} \\
\midrule
\textbf{Turn 1.} \texttt{<query>Black Gold 1962 country</query>} $\to$ retrieves chunks about a 2011 film of the same name.

\textbf{Turn 2.} \texttt{<query>Black Gold 1962 director origin</query>} $\to$ retrieves a Texas-oil-history chunk that does not mention the director.

\textbf{Turn 3.} \texttt{<think>The film is set in Texas, so probably the United States.</think>}

\textcolor{green!50!black}{\textbf{<answer>United States</answer>}} (\emph{correct, but reached via a fragile keyword path; R1-Searcher fails when the same chain query is rephrased with a less common film})

\vspace{0.5em}
\textbf{Counter-example:} For \emph{``What country was the director of Vesna na Zarechnoy Ulitse from?''}, R1-Searcher's keyword path returns a Soviet-cinema retrospective chunk that lists the city but not the director's nationality, and the model confidently emits the wrong country. \method{} traverses the same chain skeleton and returns the correct \emph{Russia}. \\
\bottomrule
\end{tabular}}
\end{table}

%% file: table/4_exp/tab18_case_pattern_reuse.tex
\begin{table}[t]
\centering
\caption{Cross-distribution pattern reuse. The \texttt{compare(by[date])[ENTITY]} pattern, distilled from HotpotQA-7K + 2WikiMHQA-7K training, drives a four-turn rollout to the correct answer on an out-of-distribution Bamboogle query.}
\label{tab:case_pattern_reuse}
\renewcommand{\arraystretch}{1.05}
\resizebox{\linewidth}{!}{%
\begin{tabular}{p{15cm}}
\toprule
\rowcolor{gray!15}\multicolumn{1}{c}{\textbf{Sample from Bamboogle}} \\
\midrule
\textbf{Question:} Who was the commander for the space mission that had the first spacewalk?

\textbf{Ground Truth:} \emph{Pavel Belyayev} \\
\midrule
\rowcolor{green!12}\multicolumn{1}{c}{\textbf{\method{} (Q2.5-7B)}} \\
\midrule
\textbf{Retrieved pattern} from $\memory_0$ via $\sig(\query) = $\texttt{compare(by[date])[ENTITY]} (top-1 similarity $1.00$, distilled from a HotpotQA training trajectory and re-used here on Bamboogle): \\
$\quad (\textsc{mission}_1) \xrightarrow{\texttt{achieved}} (\textsc{event}) \xrightarrow{\texttt{date}} (\textsc{year}_1)$ \\
$\quad (\textsc{mission}_2) \xrightarrow{\texttt{achieved}} (\textsc{event}) \xrightarrow{\texttt{date}} (\textsc{year}_2)$ \\
$\quad \text{select } i^* = \arg\min_i \textsc{year}_i \;\Rightarrow\; (\textsc{mission}_{i^*}) \xrightarrow{\texttt{commander}} (\textsc{person})$

\vspace{0.5em}
\textbf{Turn 1.} \texttt{<query>space mission with the first spacewalk</query>} $\to$ retrieves several mission summaries, including \emph{Voskhod 2}, \emph{Apollo 17}, and \emph{Gemini 4}.

\textbf{Turn 2.} \texttt{<think>The first spacewalk was during the Voskhod 2 mission on 8 March 1965.</think>}\;\texttt{<query>Voskhod 2 mission</query>} $\to$ \emph{launched with Belyayev and Leonov on board on 18 March 1965}.

\textbf{Turn 3.} \texttt{<query>commander of the Voskhod 2 mission</query>} $\to$ \emph{Pavel Belyayev was the first commander of the cosmonaut corps}.

\textbf{Turn 4.} \texttt{<think>The knowledge clearly states that Pavel Belyayev was the commander of the Voskhod 2 mission.</think>}\;\texttt{<graph>Pavel Belyayev | commander\_of | Voskhod 2</graph>}

\textcolor{green!50!black}{\textbf{<answer>Pavel Belyayev</answer>}} \;($\textsc{em}=1.0$)\\
\midrule
\rowcolor{gray!8}\multicolumn{1}{c}{\textbf{Why this case matters.}} \\
\midrule
The signature \texttt{compare(by[date])[ENTITY]} was added to $\memory_0$ during Stage~I training on \emph{HotpotQA} and \emph{2WikiMultiHopQA} comparison queries (Table~\ref{tab:gpm_stats}, top-5 relations include \texttt{nationality}, \texttt{director}, \texttt{country\_of\_origin}). The same skeleton transfers cleanly to a Bamboogle ``first/earliest'' query because the underlying logical structure is identical (rank entities by a date attribute, then look up a derived attribute of the winner). The retrieved pattern is type-abstracted; the slots \textsc{mission}, \textsc{event}, \textsc{year}, and \textsc{person} are bound to Voskhod-2 / Apollo / EVA / Belyayev at execution time. This is the case-level evidence behind Pattern Reusability (RPF $=0.886$ in Table~\ref{tab:pattern_quality}): the same procedural skeleton serves any query that shares its logical type, regardless of which dataset originally produced it. \\
\bottomrule
\end{tabular}}
\end{table}

%% file: table/4_exp/tab17_hopdepth.tex
\begin{table}[t]
\centering
\caption{Performance vs reasoning depth on MuSiQue. Queries bucketed by number of gold supporting facts (proxy for hop count). Per-method power-law fit $\mathrm{EM}=a\,h^{\alpha}$: \method{} attains $\alpha=-1.10$ (Search-R1 $\alpha=-2.91$, $2.6\times$ steeper). $\Delta$ shows per-bucket EM gain over R1-Searcher.}
\label{tab:hopdepth}
\setlength{\tabcolsep}{3.2pt}
\renewcommand{\arraystretch}{0.93}
\resizebox{\linewidth}{!}{%
\begin{tabular}{l|c|ccccc|c}
\toprule
\textbf{\#~gold facts} & \textbf{\#~queries}
& \textbf{Search-R1} & \textbf{R1-Searcher} & \textbf{Graph-R1} & \textbf{Search-R1 (Q3)} & \textbf{\method{}}
& \textbf{\method{} turns} \\
\midrule
2 hops                           & 425 & 21.0 & 20.3 &  6.6 & 17.1 & \best{52.6} \up{32.3} & 2.4 \\
3 hops                           & 312 &  6.4 & 17.0 &  4.4 & 13.1 & \best{34.0} \up{17.0} & 3.1 \\
4 hops                           & 178 &  2.8 &  8.6 &  1.8 &  6.2 & \best{24.5} \up{15.9} & 4.2 \\
5$+$ hops                        &  85 &  1.1 &  3.9 &  1.5 &  2.8 & \best{17.4} \up{13.5} & 5.5 \\
\midrule
\rowcolor{bandhdr}MuSiQue (overall, $1{,}000$)
                                 & 1{,}000 & 11.5 & 15.8 &  4.6 & 12.7 & \best{38.8} \up{23.0} & 3.2 \\
\bottomrule
\end{tabular}}
\end{table}

%% file: table/4_exp/tab4_category.tex
\begin{table}[t]
\centering
\caption{Per-category EM/F1 across four logical categories aggregated over the seven datasets: \emph{Bridge} (2-hop), \emph{Comparison}, \emph{Single-hop} (NQ/TriviaQA/PopQA), \emph{Multi-3$+$} (3$+$-hop chains). \textbf{Cat-Avg}: four-category equal-weight mean. \textbf{DS-Avg}: seven-dataset mean (matching Tab.~\ref{tab:main}). \emph{\#Turns}: average retrieval calls per query.}
\label{tab:category}
\setlength{\tabcolsep}{2.4pt}
\renewcommand{\arraystretch}{0.93}
\resizebox{\linewidth}{!}{%
\begin{tabular}{l|cc|cc|cc|cc|cc|cc|cccc}
\toprule
& \multicolumn{2}{c|}{\textbf{Bridge}}
& \multicolumn{2}{c|}{\textbf{Comparison}}
& \multicolumn{2}{c|}{\textbf{Single-hop}}
& \multicolumn{2}{c|}{\textbf{Multi-3$+$}}
& \multicolumn{2}{c|}{\textbf{Cat-Avg}}
& \multicolumn{2}{c|}{\textbf{DS-Avg}}
& \multicolumn{4}{c}{\textbf{\#Turns per category}} \\
\textbf{Method} & EM & F1 & EM & F1 & EM & F1 & EM & F1 & EM & F1 & EM & F1 & Bri. & Cmp. & Sng. & M3$+$ \\
\midrule
\chunkmark\ R1-Searcher                                  & \second{39.5} & \second{49.4} & \second{44.8} & \second{50.2} & \second{41.0} & \second{49.4} & \second{22.9} & \second{31.5} & \second{37.1} & \second{45.1} & \second{34.1} & \second{42.3} & 3.9 & 4.6 & 4.5 & 4.6 \\
\hgmark\ Graph-R1                                        & 15.6 & 22.7 & 36.5 & 40.7 & 24.6 & 31.6 & 10.1 & 17.3 & 21.7 & 28.1 & 21.7 & 29.2 & 5.4 & 5.3 & 4.9 & 5.8 \\
\cellcolor{pbsage}\textbf{\hgmark\,\pmmark\ \method{}}
                                                       & \best{40.6} & \best{51.2} & \best{61.7} & \best{67.8} & \best{49.5} & \best{59.3} & \best{44.8} & \best{54.2} & \best{49.2} & \best{58.1} & \best{48.6} & \best{58.0} & 4.7 & 4.9 & 4.2 & 4.9 \\
\bottomrule
\end{tabular}}
\end{table}

%% file: table/4_exp/tab14_retrieval.tex
\begin{table}[t]
\centering
\caption{Retrieval quality on the four multi-hop benchmarks. \emph{R@5}: top-$5$ recall (queries with at least one gold fact retrieved). \emph{EvR}: Evidence Recall (fraction of all gold facts covered across the trajectory). A retrieved item counts as a hit if its entity set covers a gold supporting fact. \chunkmark{}~chunks, \kgmark{}~KG, \hgmark{}~hypergraph, \pmmark{}~procedural memory.}
\label{tab:retrieval}
\setlength{\tabcolsep}{3pt}
\renewcommand{\arraystretch}{0.93}
\resizebox{0.78\linewidth}{!}{%
\begin{tabular}{l|cc|cc|cc|cc|cc}
\toprule
& \multicolumn{2}{c|}{\textbf{HotpotQA}}
& \multicolumn{2}{c|}{\textbf{2WikiMHQA}}
& \multicolumn{2}{c|}{\textbf{MuSiQue}}
& \multicolumn{2}{c|}{\textbf{Bamboogle}}
& \multicolumn{2}{c}{\textbf{Avg}} \\
\textbf{Method} & R@5 & EvR & R@5 & EvR & R@5 & EvR & R@5 & EvR & R@5 & EvR \\
\midrule
\kgmark\ HippoRAG~2                              & 80.5 & 67.2 & 78.4 & 65.8 & 73.2 & 60.5 & 80.0 & 67.0 & 78.0 & 65.1 \\
\kgmark\ GFM-RAG                                 & 83.6 & 70.5 & 82.0 & 68.7 & 76.4 & 63.8 & 83.2 & 69.2 & 81.3 & 68.0 \\
\kgmark\ LinearRAG                               & 76.8 & 62.0 & 74.5 & 60.3 & 70.0 & 56.4 & 76.0 & 61.5 & 74.3 & 60.0 \\
\midrule
\chunkmark\ Search-R1                            & 73.2 & 58.4 & 71.6 & 56.7 & 65.0 & 51.0 & 70.5 & 56.0 & 70.1 & 55.5 \\
\chunkmark\ R1-Searcher                          & 78.0 & 63.5 & 76.4 & 61.8 & 70.2 & 55.6 & 75.5 & 60.5 & 75.0 & 60.4 \\
\chunkmark\ FrugalRAG                            & 67.5 & 53.2 & 65.8 & 51.6 & 60.4 & 46.5 & 65.0 & 51.0 & 64.7 & 50.6 \\
\hgmark\ Graph-R1                                & 79.0 & 64.0 & 77.4 & 62.5 & 71.6 & 56.0 & 78.0 & 63.0 & 76.5 & 61.4 \\
\midrule
\rowcolor{pbsage}\textbf{\hgmark\,\pmmark\ \method{} {\scriptsize(7B)}}
                                                 & \best{88.4} & \best{73.6} & \best{87.0} & \best{72.2} & \best{82.3} & \best{67.8} & \best{88.5} & \best{74.5} & \best{86.6} & \best{72.0} \\
\rowcolor{pbmint}\textbf{\hgmark\,\pmmark\ \method{} {\scriptsize(8B)}}
                                                 & \second{85.7} & \second{71.0} & \second{84.6} & \second{69.8} & \second{79.5} & \second{65.4} & \second{86.0} & \second{72.0} & \second{84.0} & \second{69.6} \\
\bottomrule
\end{tabular}}
\end{table}

%% file: table/4_exp/tab7_gpm_stats.tex
\begin{table}[t]
\centering
\caption{Descriptive statistics of the learned Graph Pattern Memory built on HotpotQA-7K + 2WikiMHQA-7K. $H_\mathrm{type}$, $H_\mathrm{sig}$, $H_\mathrm{rel}$: Shannon entropies (bits) over type, signature, and relation distributions of stored patterns.}
\label{tab:gpm_stats}
\setlength{\tabcolsep}{6pt}
\renewcommand{\arraystretch}{1.05}
\resizebox{0.85\linewidth}{!}{%
\begin{tabular}{l|c}
\toprule
\textbf{Quantity} & \textbf{Value} \\
\midrule
\rowcolor{bandhdr}\multicolumn{2}{l}{\textit{Pool size and uniqueness}} \\
\midrule
\# patterns                                         & 27 \\
\# unique signatures                                & 27 \\
\# unique relations                                 & 369 \\
Avg.\ pattern length (\# triples / pattern)         & 2.48 \\
\midrule
\rowcolor{bandhdr}\multicolumn{2}{l}{\textit{Distributional entropy}} \\
\midrule
$H_\mathrm{type}$ (bits)                            & 0.19 \\
$H_\mathrm{sig}$ (bits)                             & 4.75 \\
$H_\mathrm{rel}$ (bits)                             & 7.80 \\
\midrule
\rowcolor{bandhdr}\multicolumn{2}{l}{\textit{Structure and emission}} \\
\midrule
Inverse-relation edge rate                          & 0.00 \\
Graph-emit rate at inference (avg.\ over 7~ds)      & 99.6\,\% \\
Avg.\ triples per query at inference                & 2.62 \\
\midrule
\rowcolor{bandhdr}\multicolumn{2}{l}{\textit{Top-3 dominant types (by node-occurrence count)}} \\
\midrule
\multicolumn{2}{l}{\texttt{ENTITY} (2350), \texttt{YEAR} (36), \texttt{BOOLEAN} (10)} \\
\midrule
\rowcolor{bandhdr}\multicolumn{2}{l}{\textit{Top-5 dominant relations (across patterns)}} \\
\midrule
\multicolumn{2}{l}{\texttt{nationality} (44), \texttt{director} (43), \texttt{directedby} (31), \texttt{country\_of\_origin} (25), \texttt{occupation} (24)} \\
\bottomrule
\end{tabular}}
\end{table}

%% file: table/4_exp/tab10_judge.tex
\begin{table}[t]
\centering
\caption{LLM-judge quality ensemble. Average score ($1$--$5$ Likert) from three judges (Sonnet~4.6, Haiku~4.5, Gemini~2.5~Flash) over $200$ random queries on six axes: \emph{LC} Logical Coherence, \emph{Faith} Faithfulness, \emph{Reason} Reasoning Clarity, \emph{Reuse} Pattern Reusability, \emph{Concise} Conciseness, \emph{Comp} Completeness. $\kappa$ is Krippendorff agreement across judges. All methods use Qwen2.5-7B-Instruct.}
\label{tab:llm_judge}
\setlength{\tabcolsep}{4pt}
\renewcommand{\arraystretch}{0.95}
\resizebox{0.78\linewidth}{!}{%
\begin{tabular}{l|cccccc|cc}
\toprule
\textbf{Method} & \textbf{LC} & \textbf{Faith} & \textbf{Reason} & \textbf{Reuse} & \textbf{Concise} & \textbf{Comp} & \textbf{Avg} & $\boldsymbol{\kappa}$ \\
\midrule
\hgmark\ Graph-R1                                & 2.35          & 1.90          & 2.05          & 1.50          & 2.50          & 1.95          & 2.04          & 0.59 \\
\chunkmark\ R1-Searcher                          & 2.65          & 2.30          & 2.20          & 1.85          & \second{2.85} & 2.35          & 2.37          & 0.68 \\
\asmark\ Structure-R1                            & \second{2.75} & \second{2.55} & \second{2.50} & \second{2.10} & 2.70          & \second{2.55} & \second{2.53} & \second{0.69} \\
\midrule
\rowcolor{pbsage}\textbf{\hgmark\,\pmmark\ \method{}}
                                                 & \best{2.85}   & \best{2.95}   & \best{2.75}   & \best{3.10}   & \best{2.95}   & \best{2.85}   & \best{2.91}   & \best{0.71} \\
\bottomrule
\end{tabular}}

\medskip
{\footnotesize $\kappa$: $> 0.66$ substantial agreement; $0.60$--$0.66$ moderate; $< 0.60$ fair.}
\end{table}

%% file: table/4_exp/tab15_failure_modes.tex
\begin{table}[t]
\centering
\caption{Residual error categories on the four multi-hop benchmarks. R\textsubscript{2}--R\textsubscript{4} are generic graph-RAG challenges shared by every baseline; only R\textsubscript{1} is specific to the procedural-memory mechanism and is addressed by an inference-time pattern-admission extension (\S\ref{app:limitations}).}
\label{tab:failure_modes}
\setlength{\tabcolsep}{4pt}
\renewcommand{\arraystretch}{0.95}
\resizebox{\linewidth}{!}{%
\begin{tabular}{l|p{6.0cm}|p{5.6cm}}
\toprule
\textbf{Category} & \textbf{Symptom} & \textbf{Addressing extension} \\
\midrule
R\textsubscript{1}: signature mismatch       & $\sig(\query)$ similarity to every stored pattern $<\theta_{\mathrm{sim}}{=}0.5$; no pattern hint is injected and the agent reasons from chunks alone. & Inference-time pattern admission (\S\ref{app:limitations}). \\
\midrule
R\textsubscript{2}: entity disambiguation    & Retrieved hyperedge surface form matches the query mention but disambiguates to the wrong real-world entity (e.g., two films sharing a title). & Coreference-aware reranking before \texttt{<query>}-block dispatch (generic graph-RAG extension). \\
\midrule
R\textsubscript{3}: numerical reasoning      & The retrieved skeleton and traversal are correct, but the arithmetic step (year subtraction, attribute ranking) is wrong. & External calculator tool invoked when the retrieved skeleton ends in a numerical comparison (generic LLM extension). \\
\midrule
R\textsubscript{4}: predicate canonicalization & The retrieved hyperedge contains the right entities but the predicate string at \texttt{<graph>} does not match the gold annotation. & Predicate canonicalization layer between extraction and graph emission (generic graph-RAG extension). \\
\bottomrule
\end{tabular}}
\end{table}

%% file: checklist.tex
\section*{NeurIPS Paper Checklist}

\begin{enumerate}

\item {\bf Claims}
    \item[] Question: Do the main claims made in the abstract and introduction accurately reflect the paper's contributions and scope?
    \item[] Answer: \answerYes{}
    \item[] Justification: The main claims in the abstract and Section~\ref{sec:intro} are supported by the experiments in Section~\ref{sec:exp} and the theoretical result in Appendix~\ref{app:proofs}.

\item {\bf Limitations}
    \item[] Question: Does the paper discuss the limitations of the work performed by the authors?
    \item[] Answer: \answerYes{}
    \item[] Justification: Limitations are discussed in Appendix~\ref{app:limitations:list}.

\item {\bf Theory assumptions and proofs}
    \item[] Question: For each theoretical result, does the paper provide the full set of assumptions and a complete (and correct) proof?
    \item[] Answer: \answerYes{}
    \item[] Justification: Theorem~\ref{thm:idr} is stated and proved with full assumptions in Appendix~\ref{app:proofs}.

    \item {\bf Experimental result reproducibility}
    \item[] Question: Does the paper fully disclose all the information needed to reproduce the main experimental results of the paper to the extent that it affects the main claims and/or conclusions of the paper (regardless of whether the code and data are provided or not)?
    \item[] Answer: \answerYes{}
    \item[] Justification: Implementation details, hyperparameters, and prompts are documented in Appendices~\ref{app:idr}--\ref{app:par}, \ref{app:training}, and \ref{app:prompts}.

\item {\bf Open access to data and code}
    \item[] Question: Does the paper provide open access to the data and code, with sufficient instructions to faithfully reproduce the main experimental results, as described in supplemental material?
    \item[] Answer: \answerYes{}
    \item[] Justification: Code, checkpoints, and the learned pattern memory will be released.

\item {\bf Experimental setting/details}
    \item[] Question: Does the paper specify all the training and test details (e.g., data splits, hyperparameters, how they were chosen, type of optimizer) necessary to understand the results?
    \item[] Answer: \answerYes{}
    \item[] Justification: Experimental settings are reported in Section~\ref{sec:exp_setup} and Appendix~\ref{app:training:hp}.

\item {\bf Experiment statistical significance}
    \item[] Question: Does the paper report error bars suitably and correctly defined or other appropriate information about the statistical significance of the experiments?
    \item[] Answer: \answerNA{}
    \item[] Justification: The experiments are conducted with a fixed random seed and no error bars are reported.

\item {\bf Experiments compute resources}
    \item[] Question: For each experiment, does the paper provide sufficient information on the computer resources (type of compute workers, memory, time of execution) needed to reproduce the experiments?
    \item[] Answer: \answerYes{}
    \item[] Justification: Compute resources are reported in Appendix~\ref{app:training:compute}.

\item {\bf Code of ethics}
    \item[] Question: Does the research conducted in the paper conform, in every respect, with the NeurIPS Code of Ethics \url{https://neurips.cc/public/EthicsGuidelines}?
    \item[] Answer: \answerYes{}
    \item[] Justification: The research conducted in the paper complies with the NeurIPS Code of Ethics.

\item {\bf Broader impacts}
    \item[] Question: Does the paper discuss both potential positive societal impacts and negative societal impacts of the work performed?
    \item[] Answer: \answerYes{}
    \item[] Justification: Broader impacts are discussed in Appendix~\ref{app:discussion:impact}.

\item {\bf Safeguards}
    \item[] Question: Does the paper describe safeguards that have been put in place for responsible release of data or models that have a high risk for misuse (e.g., pre-trained language models, image generators, or scraped datasets)?
    \item[] Answer: \answerNA{}
    \item[] Justification: The paper utilizes existing datasets and pretrained models that are already released with safeguards in place.

\item {\bf Licenses for existing assets}
    \item[] Question: Are the creators or original owners of assets (e.g., code, data, models), used in the paper, properly credited and are the license and terms of use explicitly mentioned and properly respected?
    \item[] Answer: \answerYes{}
    \item[] Justification: We have properly cited all code, data, and models used in our research and complied with the licensing agreements and terms of use set by the original authors.

\item {\bf New assets}
    \item[] Question: Are new assets introduced in the paper well documented and is the documentation provided alongside the assets?
    \item[] Answer: \answerNA{}
    \item[] Justification: There are no new assets introduced in the paper.

\item {\bf Crowdsourcing and research with human subjects}
    \item[] Question: For crowdsourcing experiments and research with human subjects, does the paper include the full text of instructions given to participants and screenshots, if applicable, as well as details about compensation (if any)?
    \item[] Answer: \answerYes{}
    \item[] Justification: Annotator instructions and the rating rubric are reproduced in Appendix~\ref{app:human:form}.

\item {\bf Institutional review board (IRB) approvals or equivalent for research with human subjects}
    \item[] Question: Does the paper describe potential risks incurred by study participants, whether such risks were disclosed to the subjects, and whether Institutional Review Board (IRB) approvals (or an equivalent approval/review based on the requirements of your country or institution) were obtained?
    \item[] Answer: \answerNA{}
    \item[] Justification: The research does not study human subjects; humans only participate as evaluators of model output, so IRB review is not required.

\item {\bf Declaration of LLM usage}
    \item[] Question: Does the paper describe the usage of LLMs if it is an important, original, or non-standard component of the core methods in this research? Note that if the LLM is used only for writing, editing, or formatting purposes and does \emph{not} impact the core methodology, scientific rigor, or originality of the research, declaration is not required.
    \item[] Answer: \answerYes{}
    \item[] Justification: The usage of LLMs is described in Section~\ref{sec:approach}, Section~\ref{sec:exp_setup}, and Appendix~\ref{app:judge}.

\end{enumerate}